\documentclass{article}

\usepackage{microtype}
\usepackage{graphicx}
\usepackage{subfigure} 
\usepackage{subcaption}
\usepackage{booktabs} 

\usepackage{hyperref}

\usepackage[preprint]{icml2026}

\usepackage[table]{xcolor}
\usepackage{graphicx}
\usepackage{enumitem}
\usepackage{array}
\usepackage{mathrsfs}
\usepackage{amsthm}
\usepackage{amsmath}
\usepackage{amsfonts}
\usepackage{multirow}
\usepackage{booktabs}
\usepackage{color}
\usepackage{arydshln}
\usepackage{makecell}
\usepackage{soul}
\usepackage{bbding}
\usepackage{colortbl}
\usepackage{amssymb}
\usepackage{mathtools}
\usepackage{tcolorbox}
\usepackage{diagbox}

\usepackage[capitalize,noabbrev]{cleveref}

\theoremstyle{plain}
\newtheorem{theorem}{Theorem}[section]
\newtheorem{proposition}[theorem]{Proposition}

\theoremstyle{definition}

\theoremstyle{remark}

\usepackage[textsize=tiny]{todonotes}

\icmltitlerunning{Beyond the Vision Encoder: Identifying and Mitigating Spatial Bias in  Large Vision-Language Models}

\begin{document}

\twocolumn[
  \icmltitle{Beyond the Vision Encoder: Identifying and Mitigating Spatial Bias in \\ Large Vision-Language Models}



  \icmlsetsymbol{equal}{*}

  \begin{icmlauthorlist}
    \icmlauthor{Yingjie Zhu}{hit,pcl}
    \icmlauthor{Xuefeng Bai}{hit}
    \icmlauthor{Kehai Chen}{hit}
    \icmlauthor{Yang Xiang}{pcl}
    \icmlauthor{Youcheng Pan}{pcl}
    \icmlauthor{Yongshuai Hou}{pcl}
    \icmlauthor{Weili Guan}{hit}
    \icmlauthor{Jun Yu}{hit}
    \icmlauthor{Min Zhang}{hit}
  \end{icmlauthorlist}

  \icmlaffiliation{hit}{Harbin Institute of Technology, Shenzhen, China}
  \icmlaffiliation{pcl}{Peng Cheng Laboratory, Shenzhen, China}

  \icmlcorrespondingauthor{Xuefeng Bai}{baixuefeng@hit.edu.cn}

  \icmlkeywords{Machine Learning, ICML}

  \vskip 0.3in
]



\printAffiliationsAndNotice{}  

\begin{abstract}
Large Vision-Language Models (LVLMs) have achieved remarkable success across a wide range of multimodal tasks, yet their robustness to spatial variations remains insufficiently understood. 
In this work, we conduct a systematic study of the spatial bias of LVLMs, examining how models respond when identical key visual information is placed at different locations within an image. Through controlled probing experiments, we observe that current LVLMs often produce inconsistent outputs under such spatial shifts, revealing a clear spatial bias in their semantic understanding.
Further analysis indicates that this bias does not stem from the vision encoder, but rather from a mismatch in attention mechanisms between the vision encoder and the large language model, which disrupts the global information flow. 
Motivated by this insight, we propose Adaptive Global Context Injection (AGCI), a lightweight mechanism that dynamically injects shared global visual context into each image token. 
AGCI works without architectural modifications, mitigating spatial bias by enhancing the semantic accessibility of image tokens while preserving the model's intrinsic capabilities.
Extensive experiments demonstrate that AGCI not only enhances the spatial robustness of LVLMs, but also achieves strong performance on various downstream tasks and hallucination benchmarks.

\end{abstract}

\section{Introduction}
Large Vision-Language Models (LVLMs) have achieved remarkable success across a wide range of multimodal tasks, including visual question answering~\citep{alayrac2022flamingo, 10769058},  image captioning~\citep{li2023blip, lu2025benchmarking}, and open-ended reasoning~\citep{zhu-etal-2025-benchmarking, wang2025multimodal}. By combining powerful vision encoders with large language models (LLMs), these systems are able to integrate information from both modalities and perform complex reasoning. Despite these advances, LVLM still exhibits fundamental limitations when it comes to spatially robust semantic understanding of visual content~\citep{imam2025can, li2025mihbench, qi2025beyond}.


Recent efforts have begun to explore the \textit{spatial bias} of LVLMs, motivated by its hypothesized connection to object hallucination. \citet{xing2024mitigating} show that the widely used Rotary Position Embedding (RoPE)~\citep{su2024roformer} introduces a long-term decay effect~\citep{peng2024yarn}, which impedes LVLMs from effectively capturing visual cues located linearly far from text tokens. 
In contrast, \citet{zhu2025mitigating} argue that such findings lack generalizability across architectures and introduce two novel attention calibration mechanisms to rectify spatially unbalanced attention with LVLMs.
While insightful, their exploration of spatial bias remains confined to the context of object hallucination—a phenomenon that may also be influenced by other confounding factors, such as the inherent hallucination tendencies of LLMs.
Consequently, a rigorous examination of spatial bias grounded in the fundamental aspect of \textit{semantic understanding} remains largely unexplored.
Moreover, their analyses focus primarily on \textit{attention distributions},
failing to delve into the underlying causes of spatial bias in LVLMs.



To fill this gap, 
this work conducts a systematic investigation into the spatial robustness of LVLMs' semantic understanding when subjected to positional variations of critical visual information. 
Specifically, we construct a probing dataset designed for image-text matching, where the same semantic content is placed in various spatial locations.
The results show that LVLMs are highly sensitive to these alterations, often producing inconsistent or even contradictory outputs under such shifts.
This phenomenon highlights a critical weakness in the spatial understanding capability of current LVLMs, indicating that their integration of image features is not position-invariant but biased toward certain preferences. 
Meanwhile, the observed spatial position preferences of LVLMs further demonstrate that the bias should not be attributed to the long-term decay property of RoPE.

To uncover the root cause of this vulnerability, 
we examine LVLMs from two stages of visual information processing: perception and semantic understanding~\citep{li2025visual}.
Our eraser search~\citep{li2016understanding,de-cao-etal-2020-decisions} experiments on LVLM's perceptual ability confirm that it consistently perceives visual features of key content, regardless of the spatial position of the key image. 
Building on this, we further analyze whether the vision encoder's semantic understanding is robust to spatial position.
The high and stable similarity between text embeddings and visual features across different locations suggest that semantic encoding is also robust to spatial variation.
These observations rule out the perceptual ability and vision encoder as the source of spatial bias, suggesting that the issue originates in the LLM portion of the LVLM, where the visual features are processed for multimodal reasoning.


Based on the above observations, we hypothesize that spatial bias stems from the \textit{mechanism mismatch} between the bidirectional self-attention used in vision encoders (typically ViT-based~\citep{dosovitskiy2021an} in LVLM) and the causal attention employed in LLMs. 
While vision encoders allow image tokens to exchange contextual information globally and symmetrically, the causal constraint in LLMs fundamentally disrupts this interaction pattern: earlier tokens cannot access subsequent contextual information as originally designed, thereby breaking the symmetrical global information flow. Consequently, during cross‑modal reasoning, the contribution of each image token becomes overly dependent on its position—manifesting as the observed spatial bias.

To address this, we propose Adaptive Global Context Injection (AGCI), a lightweight and effective mechanism that explicitly reinforces global visual semantics at the LLM stage. In specific, AGCI injects the global visual context into each image token based on semantic similarity, strengthening tokens with less information while preserving those that are already semantically aligned. This design restores stable access to global context of image during cross-modal reasoning without modifying the underlying model architecture.
We adapt AGCI for broader downstream tasks. Results on six benchmark datasets show that AGCI can enhance LVLMs not only on general VQA, but also on hallucination‑ and OCR-oriented tasks, demonstrating the effectiveness and generalizability of AGCI. And the visualization of information flow from image token to text token further emphasizes the importance of incorporating global image context.

To summarize, our contributions are threefold. 
First, we provide a systematic investigation of spatial robustness in LVLMs's semantic understanding, empirically demonstrating through novel probes that LVLMs' predictions vary significantly with the spatial location of identical visual content. 
Second, we pinpoint the root cause of this bias to the loss of global image context caused by the causal attention mechanism in LLMs, rather than to deficiencies in the vision encoder.
Third, we introduce Adaptive Global Context Injection (AGCI), a simple yet effective method that improves spatial robustness without sacrificing downstream performance, validated through both probing tasks and six multimodal benchmarks\footnote{Our data and code are available at \url{https://github.com/AAAndy-Zhu/AGCI}}.

\begin{figure*}[t]
    \centering
    \includegraphics[width=\linewidth]{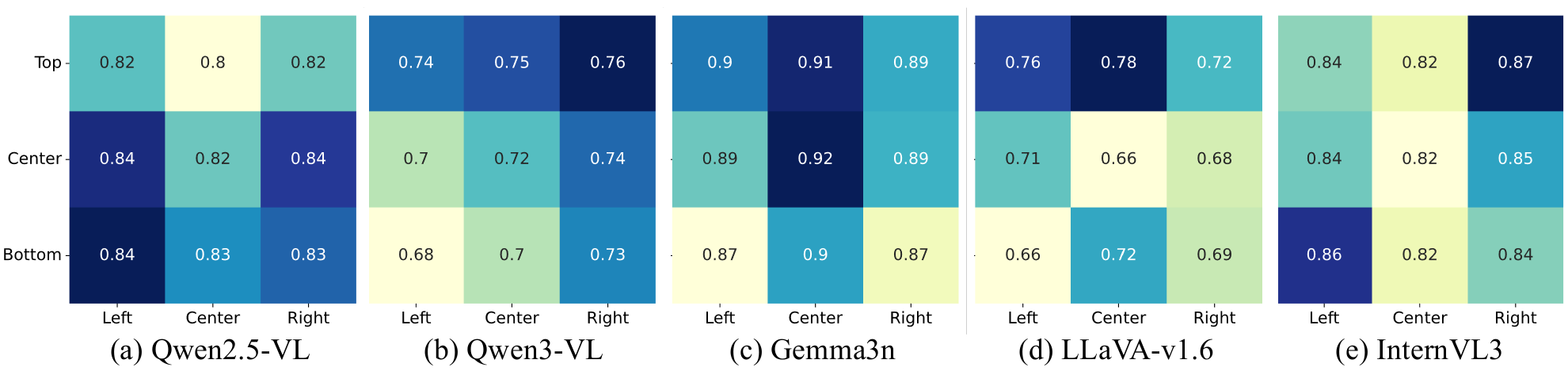}
    \caption{Results on our probe dataset, where each cell reports the model’s accuracy when the key image $I_m$ is placed at a specific grid location $n$ in the $3\times 3$ composite image $\boldsymbol{I}_{m,n}$.}
    \label{fig:prob_results}
\end{figure*}

\section{Related Work}

\textbf{Large Vision-Language Models.}
LVLMs combine both visual and textual inputs, providing a more comprehensive understanding of visual spatial relationships, objects, and scenes~\citep{bordes2024introduction}. Existing LVLMs typically comprise a visual encoder~\citep{dosovitskiy2021an,pmlr-v139-radford21a}, a projector~\citep{alayrac2022flamingo}, and a pre-trained LLM~\citep{touvron2023llama}. Through pre-training with image-text pairs and fine-tuning with preference or instruction, current LVLMs, like LLaVA~\citep{liu2023visual} and Qwen2.5-VL~\citep{bai2025qwen2}, have been successful in dialogue~\citep{zhu2025internvl3}, question answering~\citep{zhu2023minigpt}, and complex reasoning~\citep{zhu-etal-2025-benchmarking}. Nonetheless, LVLMs still exhibit numerous biases~\citep{ruggeri2023multi,wang2024vlbiasbench,zhang2025evaluating}, which diminish the trustworthiness of their response. In this paper, we systematically investigate the under-explored issue of spatial bias of LVLMs and introduce a mitigation strategy from the perspective of global image context.

\textbf{Position Bias.}  
Position bias has been widely studied in LLMs
The pioneering work reveals that LLMs typically suffer from ``the lost in the middle" phenomenon when processing multiple documents~\citep{liu-etal-2024-lost}. Similar biases have been observed in  LLM-as-a-judge scenarios, where models favor earlier responses regardless of their quality~\citep{zheng2023judging,shi-etal-2025-judging}. 
To address this issue, recent works have proposed various mitigation strategies for LLMs~\citep{hsieh-etal-2024-found,yu-etal-2025-mitigate,wang2025eliminating}. 
As LVLMs inherit the autoregressive LLM backbone, these position biases naturally extend to multimodal settings. However, existing studies on position bias in LVLMs remain limited. In \textbf{multi-image} reasoning, prior work demonstrates that LVLM predictions are highly sensitive to image ordering and introduces a training-free mitigation method based on interpolating causal and bidirectional inter-image attention~\citep{tian2025identifying}. Beyond this, several studies on multi-modal in-context learning also indirectly reveal the effects of position in LVLMs, showing that images placed at the beginning contribute minimally to model  predictions~\citep{chen2024understanding,baldassini2024makes}. In the \textbf{single-image} setting, recent efforts have begun to explore spatial bias within images, often in the context of object hallucination. \citet{xing2024mitigating} first investigate how RoPE affects object hallucination in LVLMs when the object placed at different spatial positions in the image and propose a novel position alignment method to mitigate the long-term decay in RoPE. \citet{zhu2025mitigating} then argue that such explanations lack generality and propose attention calibration mechanisms to rebalance spatial attention. 
In contrast, our work provides a principled analysis grounded in the fundamental semantic understanding task and introduce an effective mitigation strategy for spatial bias in single-image scenario.

\section{Probing Spatial Robustness in Semantic Understanding: A Controlled Study}
This section presents a controlled probing task designed to systematically evaluate the spatial robustness of LVLMs’ semantic understanding ability. Specifically, we assess whether model predictions remain consistent when key information appears in different regions of an image. We first introduce the design of our probing task, and then present experimental results to reveal the vulnerabilities of LVLMs on spatial-semantic understanding.

\subsection{Task Design}
\label{sec:prob_task_design}
To systematically evaluate the spatial robustness of LVLMs, we construct a probe dataset based on image–text matching. Specifically, we randomly sample 10,000 image–caption pairs $(I_m,C_m)$ from the LAION dataset~\citep{schuhmann2022laion}. As shown in Figure~\ref{fig:probe_dataset}, for each key image $I_m$, we first randomly retrieve 8 distractor images from LAION. We then arrange $I_m$ and the 8 distractors in a $3\times3$ grid to form a composite image $\boldsymbol{I}_{m,0}$, which is presented to the LVLM together with the caption $C_m$. The model is asked a \textit{yes-or-no} question $Q_m$ to determine whether any sub-image within the composite matches the given caption $C_m$. To probe sensitivity to spatial variation, we further construct augmented composites $\{\boldsymbol{I}_{m,1}, \boldsymbol{I}_{m,2},\dots,\boldsymbol{I}_{m,8}\}$, where the original image $I_m$ is placed at different grid locations. In each case, the same question $Q_m$ is posed to the model. The final dataset contains 90,000 samples $\{\boldsymbol{I}_{m,n},C_m,Q_m | m\in\{0,1,\dots,9,999\},n\in\{0,1,\dots,8\}\}$ in total, with each image $\boldsymbol{I}_{m,n}$ featuring a resolution of $840\times840$. By comparing outputs across these variants, we can directly assess whether LVLMs yield consistent predictions in response to positional changes of key information, while all other visual and textual factors remain unchanged. More details are available in Appendix~\ref{app:probe_task}.

\subsection{Results and Findings}
We evaluate six representative LVLMs on our probing dataset, including Qwen2.5-VL-7B~\citep{bai2025qwen2}, Qwen3-VL-8B-Instruct~\citep{Bai2025Qwen3VLTR}, Gemma3n-E4B-it~\citep{team2025gemma}, LLaVA-v1.6-Mistral-7B (LLaVA-v1.6)~\citep{liu2024improved}, and InternVL3-8B. All models are evaluated in a \textbf{zero-shot} setting, without any fine-tuning.


\textbf{Main Results.}  As shown in Figure~\ref{fig:prob_results}, all LVLMs exhibit sensitivity to the spatial variation of the key image, particularly LLaVA-v1.6, with large fluctuations in accuracy across positions. 
Among all models, Qwen2.5-VL achieves
the most consistent performance, likely benefiting from its improved MRoPE~\citep{bai2025qwen2},
although slight spatial bias can still be observed. Interestly, despite sharing a similar architecture, Qwen3-VL shows a more pronounced bias, possibly attributed to the enhanced Interleaved MRoPE failing to leverage its advantages in single-image scenarios.
Moreover, we observe that the performance is not well correlated with token distance, suggesting that spatial bias might not stem from the long-term decay property of RoPE~\citep{su2024roformer}. This observation is further supported by experiments with varying resolutions of $\boldsymbol{I}_{m,n}$, where changing the number of image tokens does not lead to notable differences in the bias pattern, reinforcing that this issue is unlikely to be caused by RoPE. The results are presented in Appendix~\ref{app:image_resolution}.

\begin{figure*}[t]
    \centering
    \subfigure[Gemma3n]{
        \includegraphics[width=0.42\textwidth]{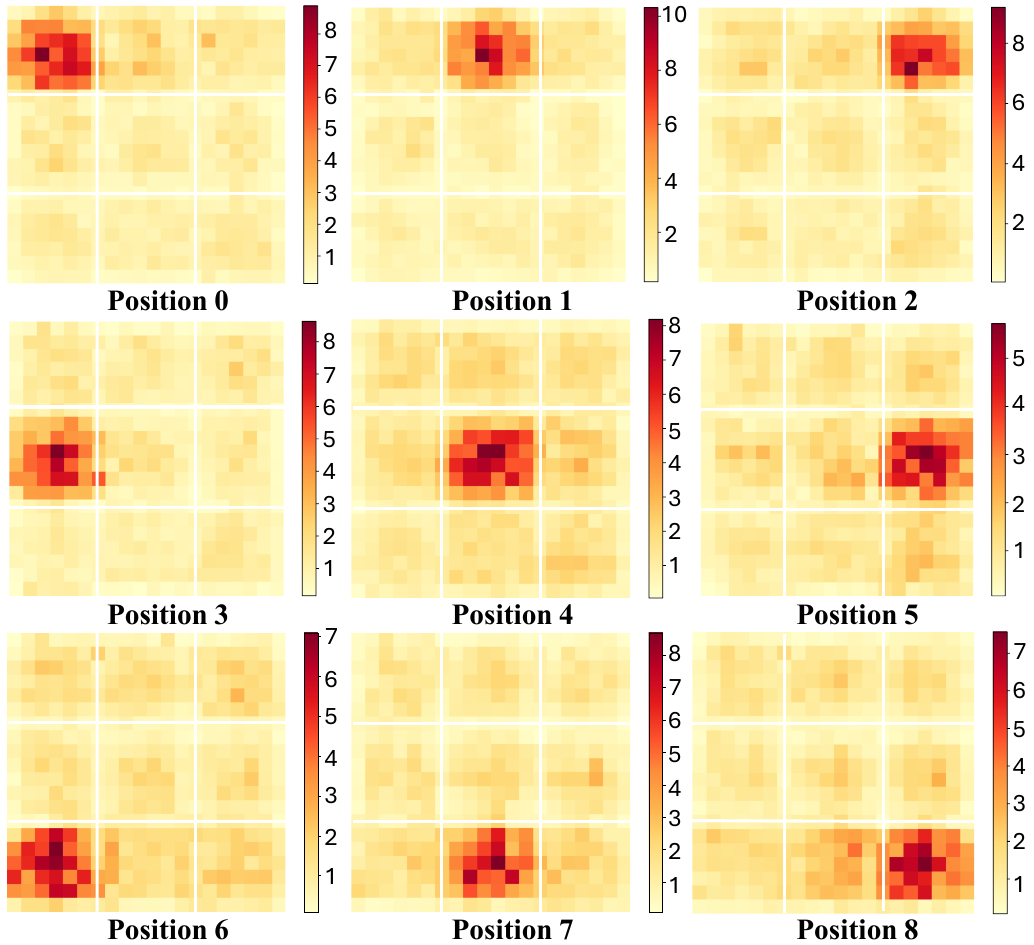}
    }
    \hspace{0.4cm}
    \subfigure[InternVL3]{
        \includegraphics[width=0.42\textwidth]{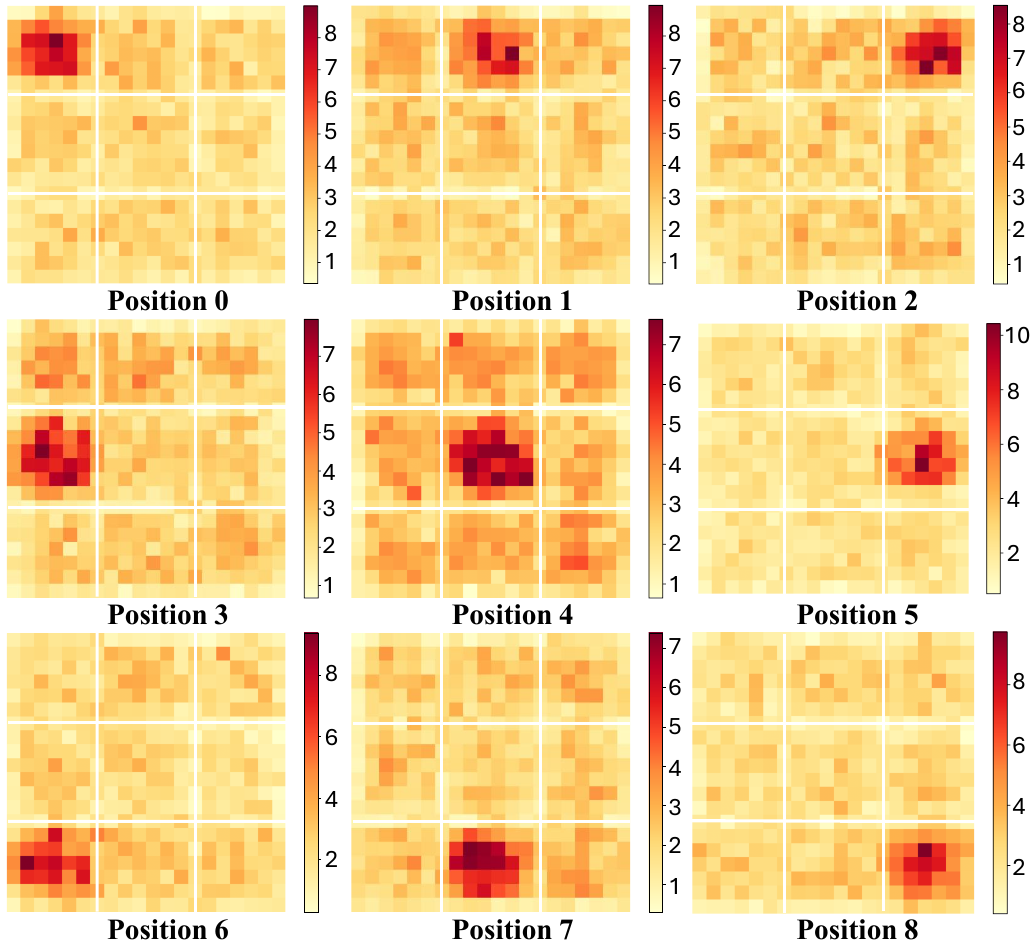}
    }
    \caption{Results on perception ability across each \textit{Position} $n$ of key image. Darker regions indicate higher importance scores, where logits change more significantly before and after masking. More results are available in Figure~\ref{fig:mask_results_more}.}
    \label{fig:mask_results}
\end{figure*}

\textbf{Impact of Model Scale.} 
We further examine the correlation between model scale and spatial bias by evaluating the Qwen2.5-VL model series on our probe dataset. As shown in Table~\ref{tab:AGCI_results}, Qwen2.5-VL-7B exhibits greater accuracy fluctuations across different grid positions ($\Delta = 1.74$), reflecting higher spatial sensitivity and stronger bias. 
With increasing model size, Qwen2.5-VL-32B and Qwen2.5-VL-72B generally achieves lower variance, which is consistent with previous observations~\citep{zhu24PromptRobust} that larger models tend to make more consistent decisions. However, the average accuracy tends to decrease, the reason can be that larger models might be more susceptible to overfitting or inefficiencies in processing specific data types.

\section{Analyzing the Origin of Spatial Bias}
To pinpoint the origin of the spatial bias in LVLMs, 
we following previous work~\citep{li2025visual} and analyze LVLMs by decomposing their visual information processing into two stages: perception and semantic understanding.

\begin{figure*}[t]
    \centering
    \includegraphics[width=\linewidth]{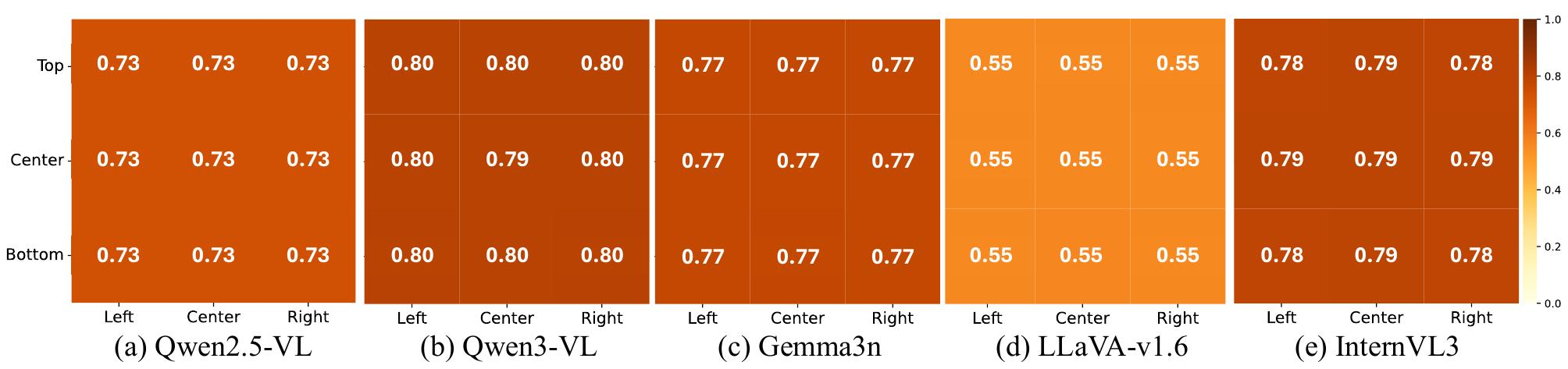}
    \caption{Understanding abilities of vision encoders across each \textit{Position} $n$. Each cell reports the similarity between visual features and text embeddings input to the LLMs when the key image $I_m$ is placed at a specific grid location $n$ in the $3\times 3$ composite image $\boldsymbol{I}_{m,n}$.}
    \label{fig:similarity_results}
\end{figure*}

\subsection{Is LVLM Perception Spatially Robust?}

\textbf{Task Design.}
\label{sec:perception_task_design}
Inspired by the work on model interpretability~\citep{si-etal-2024-denoising, zhang2025towards},
we design a set of experiments using eraser search~\citep{li2016understanding, de-cao-etal-2020-decisions} based on our probe dataset, where each region of the input image are occluded sequentially and we observe the resulting changes in the model's response. This allows us to examine how the model’s behavior is affected by different spatial locations of key image $I_m$ in the composite input $\boldsymbol{I}_{m,n}$. In specific, for each position $n$, we first randomly select 20 key images, with a total of 180 samples. As shown in Figure~\ref{fig:explore}(a), each composite image is divided into 400 non-overlapping regions, and we mask each region one at a time by perturbing its pixels to the background color (white in our experiment). Then, we calculate the difference between the logits of generated token based on the original composite image and the masked version for each LVLM and take it as the importance score of the region with respect to the response. Finally, we aggregate these importance scores into a heatmap to visualize how information from different positions within the input image influences the model's decision.

\textbf{Results and Findings.} Figure~\ref{fig:mask_results} presents the results of our masking experiments on Gemma3n and InternVL3. Both models are consistently able to identify the critical regions corresponding to the key image, irrespective of its location in the composite input.
In addition, the overall patterns remain semantically aligned with the key image and do not shift when its position changes, demonstrating that the \textbf{LVLM's perception is robust to spatial variation}. These results rule it out as the source of the spatial bias. More results are presented in Figure~\ref{fig:mask_results_more}.

\subsection{Is Vision Encoder Understanding Spatially Robust?}
\label{sec:understanding_task_design}
\textbf{Task Design.} Inspired by~\citet{pmlr-v139-radford21a}, we determine whether the vision encoder can maintain consistent semantic understanding when key image locations change by analyzing the similarity between the visual features input to the LLMs and the embeddings of the corresponding caption. 
Specifically, we first randomly select 1,000 key image-caption pairs $(I'_m, C'_m)$ from LAION dataset. As illustrated in Figure~\ref{fig:explore}(b), each image $I'_m$ is then pasted onto a white background at every position $n$, serving as synthetic image inputs $\boldsymbol{I'}_{m,n}$ for the LVLMs. We measure the semantic understanding capability of the vision encoder by evaluating the cosine similarity between the embedding representations of caption $C'_m$ and visual features of $\boldsymbol{I'}_{m,n}$ input to LLM.
\begin{equation}
    \text{Similarity} = \cos(g(f_v(\boldsymbol{I'}_{m,n})), E(C'_m)),
\end{equation}
where $g(\cdot)$ and $f_v(\cdot)$ denote the projection module and vision encoder of the LVLM, respectively. $E(C'_m)$ is the embeddings of input caption $C'_m$.

\textbf{Results and Findings.} As shown in Figure~\ref{fig:similarity_results}, the similarity scores remain highly stable regardless of the spatial placement of the key image across all five models. This consistency indicates that the vision encoder reliably extracts semantically aligned representations of the key image, independent of its spatial location in the composite. 
Moreover, combining the results in Figure~\ref{fig:prob_results}, we find a trend that LVLMs achieving higher similarity scores (e.g., Qwen2.5-VL, Gemma3n and InternVL3) also exhibit stronger and more stable performance on our probing tasks, whereas models with lower similarity scores (e.g., LLaVA-v1.6) perform worse and show greater sensitivity to positional changes. This correlation further validates that {robust semantic alignment between visual features and textual embeddings is a key factor behind reliable multimodal understanding}.

\section{Adaptive Global Context Injection (AGCI)}
Based on the above observation, we assume that the root of spatial bias may be traced to the cross-modal reasoning within the LLM backbone, where image tokens are processed under causal attention constraints. This mechanism is incompatible with the attention in the visual encoder, weakening the availability of global visual semantics during cross-modal reasoning.

\subsection{Attention Mechanism}
Attention is a core operation in both vision and language models, which dynamically allocate weights to different parts of an input sequence, allowing the model to focus on the most relevant features for a given context. Formally, 
\begin{equation}
\begin{aligned}
&\text{Attention}(\boldsymbol{Q},\boldsymbol{K},\boldsymbol{V})
= \text{softmax}\!\left(\frac{\boldsymbol{Q}\boldsymbol{K}^{\mathsf{T}}}{\sqrt{d_k}}+\boldsymbol{M}\right)\boldsymbol{V},
\end{aligned}
\end{equation}
where $\boldsymbol{Q} = \text{Pos}(\boldsymbol{W}_q\boldsymbol{X}, \boldsymbol{PE})$, $\boldsymbol{K}= \text{Pos}(\boldsymbol{W}_k\boldsymbol{X}, \boldsymbol{PE})$, and $\boldsymbol{V}= \text{Pos}(\boldsymbol{W}_v\boldsymbol{X}, \boldsymbol{PE})$ denotes the query, key and value matrices derived from the input $\boldsymbol{X}$ and the position encoding matrix $\boldsymbol{PE}$. $\boldsymbol{M}$ is the mask matrix where elements are typically set to $-\infty$ or 0 to indicate the visibility between tokens.

Current LVLMs typically adopt a Vision Transformer (ViT)~\citep{dosovitskiy2021an} as the visual encoder.
It extracts a sequence of visual tokens (patch embeddings) and applies \emph{bidirectional} self-attention (i.e., each token can attend to all other tokens in the same image), enabling global contextual aggregation of visual information.
Formally, the vision encoder uses a full attention mask
\begin{equation}
    \boldsymbol{M}_{\text{bi}} = \mathbf{0}_{L\times L}
\end{equation}
where $L$ is the number of visual tokens.

In contrast, the LLM backbone uses \emph{causal} self-attention, where each token can only attend to itself and previous tokens. The corresponding mask is
\begin{equation}
    \boldsymbol{M}_{\text{causal}}[i,j] =
    \begin{cases}
        0, & j\le i,\\
        -\infty, & j>i,
    \end{cases}
\end{equation}
which enforces a strictly \textit{lower-triangular visibility} pattern.

\begin{table*}[t]
\tiny
\centering
\caption{Results on our probe dataset where 0-8 denotes the different grid positions of key image in the composite images, $Avg$ and $\Delta$ denote the average accuracy and variance across all positions, respectively. The results with an increase after AGCI are \textbf{bolded}.}
\label{tab:AGCI_results}
\renewcommand\arraystretch{1.1}
\resizebox{0.9\linewidth}{!}{
\begin{tabular}{lccccccccccc}
\toprule
                        \textbf{\textit{Position}}       & \textbf{0} & \textbf{1} & \textbf{2} & \textbf{3} & \textbf{4} & \textbf{5} & \textbf{6} & \textbf{7} & \textbf{8} & \textbf{Avg $\uparrow$}    & $\Delta$ $\downarrow$            \\
                               \midrule
\textbf{Gemma3n-E4B}                   & 90.08      & 91.36      & 89.38      & 88.75      & 91.62      & 89.25      & 86.75      & 89.53      & 87.19      & 89.32          & 2.39          \\
 \rowcolor{blue!10} \textbf{Gemma3n-E4B-AGCI}        & 97.64      & 97.76      & 97.19      & 97.33      & 97.63      & 96.73      & 96.49      & 97.11      & 96.30       & \textbf{97.13} & \textbf{0.25} \\ \midrule
\textbf{LLaVA-v1.6-7B}                & 75.71      & 77.75      & 71.67      & 70.98      & 65.87      & 68.35      & 65.87      & 72.40      & 68.81      & 70.82          & 14.93         \\
 \rowcolor{blue!10}\textbf{LLaVA-v1.6-7B-AGCI}     & 84.84      & 85.08      & 80.90       & 80.52      & 76.86      & 77.77      & 79.19      & 83.07      & 78.49      & \textbf{80.75} & \textbf{8.05} \\\midrule
\textbf{InternVL3-8B}                 & 83.59      & 82.42      & 86.96      & 83.89      & 81.79      & 84.70       & 86.41      & 82.37      & 83.67      & 83.98          & 2.82         \\
 \rowcolor{blue!10}\textbf{InternVL3-8B-AGCI}      & 85.72      & 83.58      & 87.79      & 86.05      & 82.94      & 85.88      & 87.89      & 84.35      & 85.32      & \textbf{85.50}  & \textbf{2.56} \\\midrule
\textbf{Qwen3-VL-8B}               & 73.79      & 74.55      & 76.24      & 70.24      & 71.78      & 73.96      & 67.97      & 70.28      & 73.39      & 72.47          & 5.98          \\
 \rowcolor{blue!10}\textbf{Qwen3-VL-8B-AGCI}       & 79.50      & 80.01      & 81.13      & 76.88      & 77.68      & 79.55      & 75.44      & 76.09      & 79.64      & \textbf{87.44} & \textbf{3.45} \\\midrule
\textbf{Qwen2.5-VL-7B}             & 81.86      & 80.00      & 81.67      & 83.81      & 81.71      & 83.58      & 84.15      & 82.62      & 83.05      & 82.49          & 1.54         \\
 \rowcolor{blue!10}\textbf{Qwen2.5-VL-7B-AGCI}     & 89.25      & 87.98      & 89.29      & 90.20       & 88.30       & 90.06      & 91.04      & 89.73      & 90.23      & \textbf{89.56} & \textbf{0.84} \\
\textbf{Qwen2.5-VL-32B}            & 82.25      & 81.82      & 82.58      & 82.39      & 81.47      & 81.66      & 82.52      & 81.92      & 81.00      & 81.96          & 0.25          \\
 \rowcolor{blue!10}\textbf{Qwen2.5-VL-32B-AGCI} & 83.58      & 83.23      & 83.69      & 83.48      & 82.67      & 82.68      & 83.62      & 83.20      & 81.93      & \textbf{83.12} & 0.31          \\
\textbf{Qwen2.5-VL-72B}            & 70.33      & 70.61      & 71.64      & 69.93      & 70.39      & 70.84      & 68.60      & 68.40      & 70.36      & 70.12          & 0.95          \\
 \rowcolor{blue!10}\textbf{Qwen2.5-VL-72B-AGCI} & 86.11      & 85.98      & 86.62      & 85.88      & 85.73      & 86.00      & 86.08      & 85.80      & 86.79      & \textbf{86.11} & \textbf{0.12} \\
\bottomrule
\end{tabular}}
\end{table*}

\subsection{The Proposed Method}

In traditional LVLMs, the unidirectional constraint in LLM differs fundamentally from the bidirectional attention within vison encoder. It may hinder global context exchange among image tokens during cross-modal processing, causing image tokens to gradually lose access to holistic image information as they are propagated through the Transformer layers.

To mitigate this issue, we propose Adaptive Global Context Injection (AGCI), which injects shared global image context into each image token in an adaptive manner.
Given the image token sequence $\{\tilde{v}_i\}_{i=1}^{L}$, we first compute the mean representation as   global image context 
\begin{equation}
    \tilde{g} = \frac{1}{L}\sum_{i=1}^{L} \tilde{v}_i.
\end{equation}
Then, for each image token $\tilde{v}_i$, we measure its semantic similarity to the global context by
\begin{equation}
    w_i = \cos(\tilde{v}_i, \tilde{g}).
\end{equation}
Finally, the global context is injected into each image token with an adaptive strength proportional to $(1-w_i)$:
\begin{equation}
    \tilde{v}_i' = \tilde{v}_i + \lambda(1-w_i)\,\tilde{g}, \qquad i=1,\dots,L.
\end{equation}
where $\lambda$ is a hyperparameter to cap the maximum amount of injected global information, serving as a safety valve to better preserve fine-grained visual details.
This design ensures the tokens with lower global semantic relevance receive stronger global-context supplementation, helping preserve holistic image information during cross-modal interaction.
We provide a theoretical analysis of AGCI from the perspective of mutual information in Appendix~\ref{app:mi_agci}.

\begin{table*}[t]
\tiny
\centering
\caption{Results on downstream tasks. The best results are \textbf{bolded}.}
\label{tab:downstream}
\renewcommand\arraystretch{1.1}
\resizebox{0.8\linewidth}{!}{
\begin{tabular}{lcccccc}
\toprule
\multirow{2.3}{*}{\textbf{Models}} & \multicolumn{2}{c}{\textbf{$\text{MMMU-Pro}_\text{direct}$}}                                          & \multirow{2.3}{*}{\textbf{$\text{ScienceQA}_\text{image}$}} & \multirow{2.3}{*}{\textbf{$\text{CRPE}_\text{relation}$}} & \multirow{2.3}{*}{\textbf{HallusionBench}}
& \multirow{2.3}{*}{\textbf{TextVQA}}\\
\cmidrule(lr){2-3}
                                 & \multicolumn{1}{l}{\textbf{4-option}} & \multicolumn{1}{l}{\textbf{10-option}} &                                     &                                &            &                              \\
                                 \midrule
\textbf{Gemma3n}               & 36.53                      & 23.85                     & \textbf{83.94}              & 69.22          & \textbf{66.04}                   & 68.97            \\
\rowcolor{blue!10}\textbf{Gemma3n-AGCI}    & \textbf{41.48}             & \textbf{28.37}            & 83.59              & \textbf{69.97} & 63.04                   & \textbf{70.54}   \\ \hdashline
\textbf{LLaVA-v1.6}            & 33.90                       & \textbf{20.15}                     & 72.88              & 68.84          & 53.63                   & 63.85            \\
\rowcolor{blue!10}\textbf{LLaVA-v1.6-AGCI} & \textbf{34.21}             & 20.09                     & \textbf{73.33}     & \textbf{68.86} & \textbf{54.47}          & \textbf{65.68}   \\\hdashline
\textbf{InternVL3}             & 53.17                      & \textbf{37.92}                     & 98.22              & \textbf{76.15}          & 65.09                   & 81.12            \\
\rowcolor{blue!10}\textbf{InternVL3-AGCI}  & \textbf{53.35}             & 37.60                      & \textbf{98.41}     & 75.63          & \textbf{66.14}          & \textbf{81.81}   \\\hdashline
\textbf{Qwen3-VL}           & 38.42                      & 27.93                     & \textbf{94.79}              & 77.68          & \textbf{73.50}                    & 80.34            \\
\rowcolor{blue!10}\textbf{Qwen3-VL-AGCI}   & \textbf{45.45}             & \textbf{34.78}            & 94.05              & \textbf{77.92} & 72.56                   & \textbf{81.13}   \\\hdashline
\textbf{Qwen2.5-VL}         & 41.18                      & 28.81                     & 88.60               & 76.67          & \textbf{71.08}                  & 77.72            \\
\rowcolor{blue!10}\textbf{Qwen2.5-VL-AGCI} & \textbf{44.57}             & \textbf{31.64}            & \textbf{88.84}     & \textbf{76.95} & 69.40                    & \textbf{82.65}   \\
\bottomrule
\end{tabular} }
\end{table*}

\section{Experiments}

\subsection{Experimental Settings}
In this section, we evaluate the effectiveness of the proposed AGCI on our probe dataset and six downstream multimodal tasks. Since the training data of mainstream LVLMs is not publicly available, retraining these models with AGCI is infeasible. Thus, we perform lightweight LoRA~\citep{hu2022lora} fine-tuning with 10K instruct-tuning data from LLaVA~\citep{liu2023visual} to adapt each model to the modified image token representation. For LLaVA-v1.6, we set $\lambda$ to $0.3$, while all other LVLMs are set to $0.5$. Following~\citep{liu2024improved}, we only retained samples that contain \textit{exactly one image} in the input for each dataset during the evaluation on downstream tasks. We test the same five representative LVLMs as above under a zero-shot setting. More details are available in Appendix~\ref{app:implementation}.

\subsection{Results on Probing Task}

Table~\ref{tab:AGCI_results} reports the performance of each LVLM and their AGCI-enhanced variants on our probing task, where the key image is placed at different grid positions (0–8). For each model, we further report the average accuracy (Avg) and variance ($\Delta$) across all positions. The results highlight three major findings.

For \textbf{average accuracy}, AGCI consistently improves performance across all evaluated LVLMs, with particularly large gains upon previously underperforming models (e.g., LLaVA-v1.6 and Qwen3-VL). This suggests that the asymmetric interaction of image tokens caused by causal attention hinders the full utilization of visual information, and
injecting shared global visual context improves the semantic accessibility of image tokens and strengthens cross-modal evidence aggregation.

For \textbf{spatial robustness}, AGCI effectively reduces the accuracy variance across different positions for almost all models, indicating improved invariance to spatial shifts of critical visual information. Notably, LLaVA-v1.6, which originally suffers from severe instability, exhibits substantial variance reductions after applying AGCI.  This confirms that the cause of spatial bias lies in the disruption of global information flow within the LLM backbone, and preserving such information effectively alleviates this inconsistency.

For \textbf{model scale}, as the model parameters increase from 7B to 72B, AGCI delivers a more pronounced overall accuracy improvement while maintaining low variance, particularly for Qwen2.5-VL-72B, where accuracy surges from 70.12 to 86.11, with variance also dropping from 0.95 to 0.12. This indicates that spatial bias is not merely a small-model artifact, and explicitly reinforcing global context can further stabilize cross-position performance even for larger LVLMs.

\subsection{Results on Downstream Tasks}
\label{sec:downstream_test}

\textbf{MMMU-Pro.} We first evaluate the proposed method on MMMU-Pro~\citep{yue-etal-2025-mmmu}, which is an enhanced multimodal benchmark designed to rigorously assess the true understanding capabilities of LVLMs. 
As shown in Table~\ref{tab:downstream}, AGCI improves the performance of almost all models when confronted with such complex reasoning tasks, with particularly notable gains on both settings. 
This indicates that mitigating spatial bias benefits complex reasoning scenarios where models must integrate multiple visual cues and maintain a coherent global understanding of the image. The larger improvements observed on the 10-option setting further suggest that AGCI enhances the robustness of fine-grained reasoning under increased decision complexity.

\textbf{ScienceQA.} We then evaluate our method on the ScienceQA dataset~\citep{lu2022learn}, which is collected from elementary and high school science curricula. 
The results in Table~\ref{tab:downstream} show that AGCI can consistently improve or maintain LVLMs' performance on science-related tasks. 
These results suggest that injecting global visual semantics helps the LVLM better align visual evidence with textual knowledge, leading to better integration of visual and textual information for precise knowledge-based reasoning.

\textbf{Hallucination Benchmark.} We also evaluate AGCI on two comprehensive multimodal hallucination benchmarks, i.e., CRPE~\citep{wang2024all} and HallusionBench~\citep{guan2024hallusionbench}.  According to Table~\ref{tab:downstream}, we find that AGCI generally reduces hallucination tendencies and improves the prediction reliability of most LVLMs. For instance, AGCI improves the performance of InternVL3 on HallusionBench from 65.09 to 66.14, while other models achieve a small gain on CRPE. Although some models, such as Gemma3n, exhibit minor drops, the overall trend suggests that AGCI helps alleviate the negative effects of spatial bias, leading to more stable and trustworthy predictions.

\textbf{TextVQA.} TextVQA~\citep{Singh_2019_CVPR} evaluates LVLM's ability to recognize and reason over scene text, which is highly sensitive to spatial localization. Hence, we further validate the effectiveness of our method on this task. Based on Table~\ref{tab:downstream}, we can observe that AGCI consistently improves performance across all models, with especially strong gains for  Qwen2.5-VL (from 77.72 to 82.65) and LLaVA-v1.6 (+4.93). This suggests that restoring global visual semantics helps the LLM better aggregate dispersed textual regions in the image, alleviating spatial bias and improving the alignment between textual queries and relevant OCR tokens.

\begin{table}[t]
\tiny
   \centering
\caption{Accuracy on POPE MSCOCO dataset, \textit{Avg.} indicates the average accuracy. The best results are \textbf{bolded}. The results of baselines based on LLaVA-v1.5 are reported by~\citet{xing2024mitigating}. The complete results are presented in Table~\ref{tab:pope1}.}
\label{tab:pope}
\renewcommand\arraystretch{1.1}
\resizebox{0.9\linewidth}{!}{
\begin{tabular}{lcccc}
\toprule
\textbf{Models}                & \textbf{Adv.}  & \textbf{Pop.}  & \textbf{Rand.} & \textbf{Avg.}  \\
\midrule
\textbf{Gemma3n}               & 86.77          & 87.90          & 88.93          & 87.87          \\
\rowcolor{blue!10}\textbf{Gemma3n-AGCI}    & \textbf{87.07} & \textbf{88.37} & \textbf{89.40} & \textbf{88.28} \\
\hdashline
\textbf{LLaVA-v1.6}            & 87.10          & 88.70          & 90.80          & 88.87          \\
\rowcolor{blue!10}\textbf{LLaVA-v1.6-AGCI} & \textbf{87.17} & \textbf{89.23} & \textbf{91.40} & \textbf{89.27} \\
\hdashline
\textbf{InternVL3}             & 86.43          & 88.00          & 89.20          & 87.88          \\
\rowcolor{blue!10}\textbf{InternVL3-AGCI}  & \textbf{86.57} & \textbf{88.13} & \textbf{89.57} & \textbf{88.09} \\
\hdashline
\textbf{Qwen3-VL}           & 82.89          & 84.10          & 86.54          & 84.51          \\
\rowcolor{blue!10}\textbf{Qwen3-VL-AGCI}   & \textbf{84.67} & \textbf{86.57} & \textbf{89.63} & \textbf{86.96} \\
\hdashline
\textbf{Qwen2.5-VL}         & 87.77          & 90.67          & 94.13          & 90.86          \\
\rowcolor{blue!10}\textbf{Qwen2.5-VL-AGCI} & \textbf{89.00} & \textbf{90.73} & \textbf{93.17} & \textbf{90.97} \\
\hdashline
\textbf{LLaVA-v1.5}            & 78.96          & 81.88          & 83.29 & 81.38 \\
\textbf{LLaVA-v1.5-VCD}        & 80.88 & 85.38          & 87.73          & 84.66          \\
\textbf{LLaVA-v1.5-RLHF}        & 82.60 & 83.90          & 85.90          & 84.13          \\
\textbf{LLaVA-v1.5-CCA}        & \textbf{85.67} & \textbf{86.87}          & 88.03          & 86.86          \\
\rowcolor{blue!10}\textbf{LLaVA-v1.5-AGCI} & 85.50          & \textbf{86.87} & \textbf{88.57} & \textbf{86.98} \\
\bottomrule
\end{tabular}}
\end{table}

\textbf{POPE.} POPE~\citep{li-etal-2023-evaluating} provides a detailed evaluation of object hallucination in LVLMs. To demonstrate the benefits of AGCI, we evaluate our method with POPE built on the validation set of MSCOCO~\citep{10.1007/978-3-319-10602-1_48} under adversarial (\textit{Adv.}), popular (\textit{Pop.}), and random (\textit{Rand.}) negative sampling strategies. We additionally train LLaVA-v1.5~\citep{liu2024improved} with AGCI from scratch to compare three baseline methods: VCD~\citep{leng2024mitigating}, RLHF~\citep{sun-etal-2024-aligning}, and CCA~\citep{xing2024mitigating}. As shown in Table 3, AGCI consistently improves performance on POPE across different negative sampling strategies, indicating enhanced visual grounding and reduced hallucination. Notably, Qwen3-VL benefits substantially from AGCI, with its average accuracy improving from 84.51 to 86.96, showing stronger robustness under adversarial and popular sampling. Similar improvements are observed for LLaVA-v1.6 and InternVL3, both achieving higher average accuracy after applying AGCI. 
Meanwhile, for LLaVA-v1.5, our AGCI still consistently outperforms  baselines across most evaluation settings, even though our method is not specifically designed for the object hallucination task.
These results suggest that injecting global visual semantics effectively reduces reliance on language priors and improves model's robustness across diverse LVLM architectures.


\begin{figure}[t]
    \centering
    \includegraphics[width=0.9\linewidth]{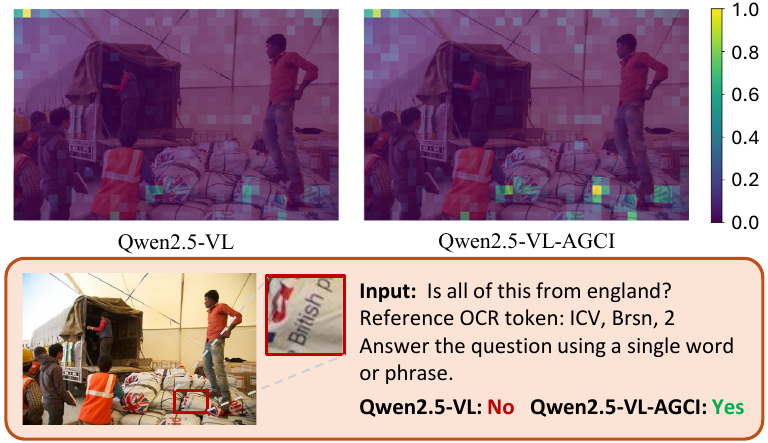}
    \caption{An example of visualization of information flow from image token to text token. More cases are provided in Appendix~\ref{app:case_study}.}
    \label{fig:case_study}
\end{figure}


\subsection{Analysis of Information Flow}
\label{sec:information_flow}
To analyze the effect of AGCI on cross-modal interactions, we present a case study from TextVQA, with the visualization of information flow in Qwen2.5-VL and the AGCI-finetuned version in Figure~\ref{fig:case_study}. 
The baseline Qwen2.5-VL exhibits diffused and weak attention over the image, where the highest attention is incorrectly concentrated on the upper-left region of the image, which is irrelevant to the question, leading to an incorrect prediction. By contrast, after applying AGCI, attention becomes more concentrated on the regions containing the key textual cues, especially for the region containing the word \textbf{“British”} (the brightest patch), which serves as the critical OCR cue for answering the question.
By injecting global context into image tokens, AGCI reduces the dominance of position-dependent local cues and provides a more balanced contextual representation, enabling the model to better identify and attend to task-relevant regions. 

\begin{figure}[t]
    \centering
    \includegraphics[width=\linewidth]{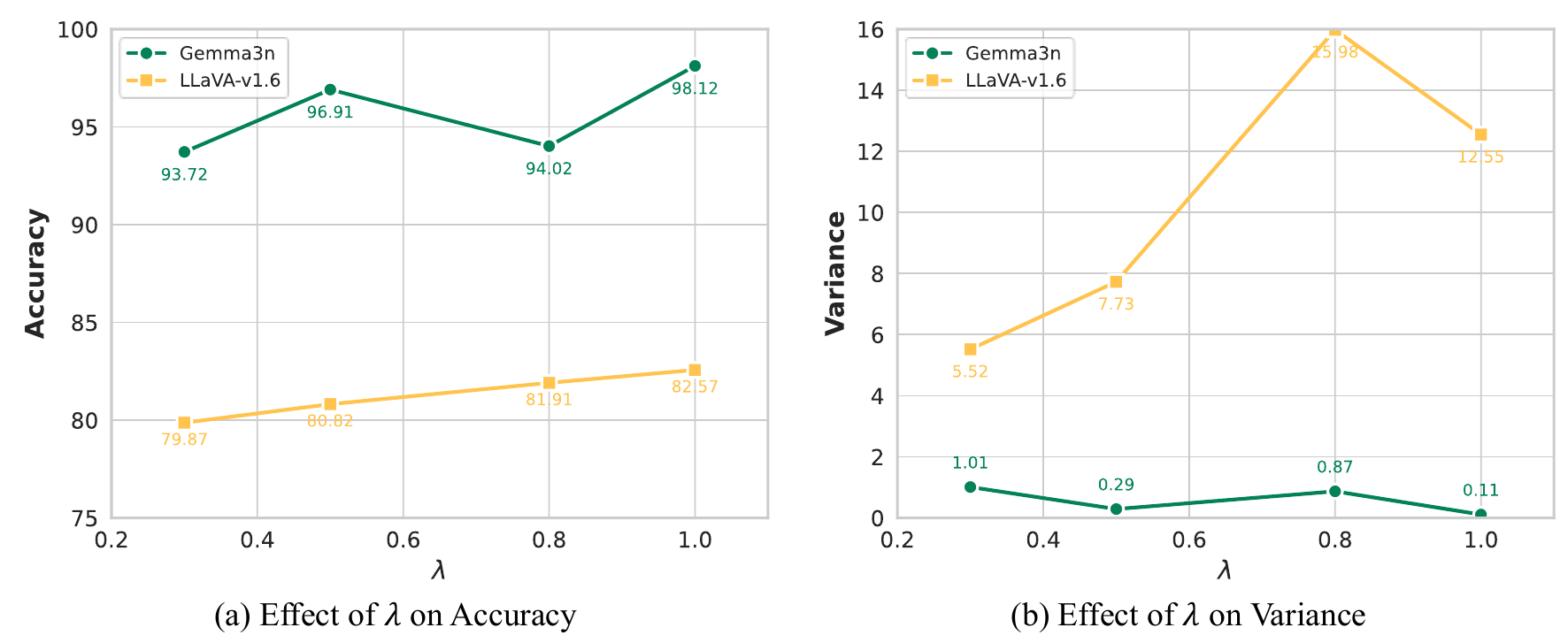}
    \caption{Effect of $\lambda$ on the Probe Dataset.}
    \label{fig:lambda}
\end{figure}

\subsection{Impact of $\lambda$}
To study the influence of the injection strength in AGCI, we vary the injection weight $\lambda$ and evaluate model performance on 9K probe samples from our dataset. As shown in Figure~\ref{fig:lambda}, Gemma3n exhibits consistently strong performance across different $\lambda$, achieving high accuracy while maintaining very low variance. As $\lambda$ increases, its overall performance shows a gradual improvement, accompanied by a decreasing trend in variance, suggesting that for robust LVLMs, increasing global context injection within a reasonable range can further enhance cross-modal information integration while preserving stable and reliable spatial reasoning. For LLaVA-v1.6, increasing $\lambda$ leads to a steady improvement in accuracy, indicating that AGCI enables the weaker model to incorporate richer visual information during cross-modal interaction, thereby supporting more accurate understanding and reasoning. However, the variance exhibits a non-monotonic trend, peaking at intermediate values of $\lambda$. We assume that for weaker LVLMs with lower baseline performance, smaller values of $\lambda$ are preferable, as excessive global context injection may overly perturb token representations, leading to increased instability and degraded robustness. 

\section{Conclusion}
This work presents a systematic investigation of spatial bias in LVLMs, showing that inconsistent predictions under spatial shifts originate from the disruption of global information flow in the LLM. To address this issue, we propose Adaptive Global Context Injection (AGCI), which injects the global visual context into each image token based on semantic similarity. Experiments demonstrate that AGCI not only enhances the spatial robustness but also improves LVLMs' performance across multimodal benchmarks with lightweight fine-tuning. Further analysis on cross modal information flow emphasizes the importance of incorporating global visual context.

\section*{Impact Statements}

This paper presents work whose goal is to elucidate and advance the spatial robustness of large vision-language models. There are many potential societal consequences of our work, none of which we feel must be specifically highlighted here.

\bibliography{example_paper}

\begin{thebibliography}{53}
\providecommand{\natexlab}[1]{#1}
\providecommand{\url}[1]{\texttt{#1}}
\expandafter\ifx\csname urlstyle\endcsname\relax
  \providecommand{\doi}[1]{doi: #1}\else
  \providecommand{\doi}{doi: \begingroup \urlstyle{rm}\Url}\fi

\bibitem[Alayrac et~al.(2022)Alayrac, Donahue, Luc, Miech, Barr, Hasson, Lenc, Mensch, Millican, Reynolds, et~al.]{alayrac2022flamingo}
Alayrac, J.-B., Donahue, J., Luc, P., Miech, A., Barr, I., Hasson, Y., Lenc, K., Mensch, A., Millican, K., Reynolds, M., et~al.
\newblock Flamingo: a visual language model for few-shot learning.
\newblock \emph{Advances in neural information processing systems}, 35:\penalty0 23716--23736, 2022.

\bibitem[Bai et~al.(2025{\natexlab{a}})Bai, Cai, Chen, Chen, Chen, Cheng, Deng, Ding, Fang, Gao, Ge, Ge, Guo, Huang, Huang, Huang, Hui, Jiang, Li, Li, Li, Li, Lin, Lin, Liu, Liu, Liu, Liu, Liu, Liu, Lu, Luo, Lv, Men, Meng, Ren, yi~Ren, Song, Sun, Tang, Tu, Wan, Wang, Wang, Wang, Wang, Xie, Xu, Xu, Xu, Yang, Yang, Yang, Yang, Yu, Zhang, Zhang, Zhang, Zheng, Zhong, Zhou, Zhou, Zhou, Zhu, and Zhu]{Bai2025Qwen3VLTR}
Bai, S., Cai, Y., Chen, R., Chen, K., Chen, X.-H., Cheng, Z., Deng, L., Ding, W., Fang, R., Gao, C., Ge, C., Ge, W., Guo, Z., Huang, Q., Huang, J., Huang, F., Hui, B., Jiang, S., Li, Z., Li, M., Li, M., Li, K., Lin, Z., Lin, J., Liu, X., Liu, J., Liu, C., Liu, Y., Liu, D., Liu, S., Lu, D., Luo, R., Lv, C., Men, R., Meng, L.~Y., Ren, X., yi~Ren, X., Song, S., Sun, Y.-C., Tang, J., Tu, J., Wan, J., Wang, P., Wang, P., Wang, Q., Wang, Y., Xie, T., Xu, Y., Xu, H., Xu, J., Yang, Z., Yang, M., Yang, J., Yang, A., Yu, B., Zhang, F., Zhang, H., Zhang, X., Zheng, B., Zhong, H., Zhou, J., Zhou, F., Zhou, J., Zhu, Y., and Zhu, K.
\newblock Qwen3-vl technical report.
\newblock \emph{arXiv preprint arXiv:2511.21631}, 2025{\natexlab{a}}.

\bibitem[Bai et~al.(2025{\natexlab{b}})Bai, Chen, Liu, Wang, Ge, Song, Dang, Wang, Wang, Tang, et~al.]{bai2025qwen2}
Bai, S., Chen, K., Liu, X., Wang, J., Ge, W., Song, S., Dang, K., Wang, P., Wang, S., Tang, J., et~al.
\newblock Qwen2. 5-vl technical report.
\newblock \emph{arXiv preprint arXiv:2502.13923}, 2025{\natexlab{b}}.

\bibitem[Baldassini et~al.(2024)Baldassini, Shukor, Cord, Soulier, and Piwowarski]{baldassini2024makes}
Baldassini, F.~B., Shukor, M., Cord, M., Soulier, L., and Piwowarski, B.
\newblock What makes multimodal in-context learning work?
\newblock In \emph{Proceedings of the IEEE/CVF Conference on Computer Vision and Pattern Recognition}, pp.\  1539--1550, 2024.

\bibitem[Bordes et~al.(2024)Bordes, Pang, Ajay, Li, Bardes, Petryk, Ma{\~n}as, Lin, Mahmoud, Jayaraman, et~al.]{bordes2024introduction}
Bordes, F., Pang, R.~Y., Ajay, A., Li, A.~C., Bardes, A., Petryk, S., Ma{\~n}as, O., Lin, Z., Mahmoud, A., Jayaraman, B., et~al.
\newblock An introduction to vision-language modeling.
\newblock \emph{arXiv preprint arXiv:2405.17247}, 2024.

\bibitem[Chen et~al.(2024)Chen, Han, He, Buckley, Torr, Tresp, and Gu]{chen2024understanding}
Chen, S., Han, Z., He, B., Buckley, M., Torr, P., Tresp, V., and Gu, J.
\newblock Understanding and improving in-context learning on vision-language models.
\newblock In \emph{ICLR 2024 Workshop on Mathematical and Empirical Understanding of Foundation Models}, 2024.
\newblock URL \url{https://openreview.net/forum?id=SB2sWF3oCw}.

\bibitem[De~Cao et~al.(2020)De~Cao, Schlichtkrull, Aziz, and Titov]{de-cao-etal-2020-decisions}
De~Cao, N., Schlichtkrull, M.~S., Aziz, W., and Titov, I.
\newblock How do decisions emerge across layers in neural models? interpretation with differentiable masking.
\newblock In \emph{Proceedings of the 2020 Conference on Empirical Methods in Natural Language Processing (EMNLP)}, pp.\  3243--3255, 2020.

\bibitem[Dosovitskiy et~al.(2021)Dosovitskiy, Beyer, Kolesnikov, Weissenborn, Zhai, Unterthiner, Dehghani, Minderer, Heigold, Gelly, Uszkoreit, and Houlsby]{dosovitskiy2021an}
Dosovitskiy, A., Beyer, L., Kolesnikov, A., Weissenborn, D., Zhai, X., Unterthiner, T., Dehghani, M., Minderer, M., Heigold, G., Gelly, S., Uszkoreit, J., and Houlsby, N.
\newblock An image is worth 16x16 words: Transformers for image recognition at scale.
\newblock In \emph{International Conference on Learning Representations}, 2021.

\bibitem[Guan et~al.(2024)Guan, Liu, Wu, Xian, Li, Liu, Wang, Chen, Huang, Yacoob, et~al.]{guan2024hallusionbench}
Guan, T., Liu, F., Wu, X., Xian, R., Li, Z., Liu, X., Wang, X., Chen, L., Huang, F., Yacoob, Y., et~al.
\newblock Hallusionbench: an advanced diagnostic suite for entangled language hallucination and visual illusion in large vision-language models.
\newblock In \emph{Proceedings of the IEEE/CVF Conference on Computer Vision and Pattern Recognition}, pp.\  14375--14385, 2024.

\bibitem[Hsieh et~al.(2024)Hsieh, Chuang, Li, Wang, Le, Kumar, Glass, Ratner, Lee, Krishna, and Pfister]{hsieh-etal-2024-found}
Hsieh, C.-Y., Chuang, Y.-S., Li, C.-L., Wang, Z., Le, L., Kumar, A., Glass, J., Ratner, A., Lee, C.-Y., Krishna, R., and Pfister, T.
\newblock Found in the middle: Calibrating positional attention bias improves long context utilization.
\newblock In \emph{Findings of the Association for Computational Linguistics: ACL 2024}, pp.\  14982--14995, 2024.

\bibitem[Hu et~al.(2022)Hu, yelong shen, Wallis, Allen-Zhu, Li, Wang, Wang, and Chen]{hu2022lora}
Hu, E.~J., yelong shen, Wallis, P., Allen-Zhu, Z., Li, Y., Wang, S., Wang, L., and Chen, W.
\newblock Lo{RA}: Low-rank adaptation of large language models.
\newblock In \emph{International Conference on Learning Representations}, 2022.

\bibitem[Imam et~al.(2025)Imam, Lyu, and Aji]{imam2025can}
Imam, M.~F., Lyu, C., and Aji, A.~F.
\newblock Can multimodal llms do visual temporal understanding and reasoning? the answer is no!
\newblock \emph{arXiv preprint arXiv:2501.10674}, 2025.

\bibitem[Leng et~al.(2024)Leng, Zhang, Chen, Li, Lu, Miao, and Bing]{leng2024mitigating}
Leng, S., Zhang, H., Chen, G., Li, X., Lu, S., Miao, C., and Bing, L.
\newblock Mitigating object hallucinations in large vision-language models through visual contrastive decoding.
\newblock In \emph{Proceedings of the IEEE/CVF Conference on Computer Vision and Pattern Recognition}, pp.\  13872--13882, 2024.

\bibitem[Li et~al.(2025{\natexlab{a}})Li, Zhang, Ding, Li, and Zhang]{li2025visual}
Li, H., Zhang, Y., Ding, J., Li, Q., and Zhang, P.
\newblock Visual room 2.0: Seeing is not understanding for mllms.
\newblock \emph{arXiv preprint arXiv:2511.12928}, 2025{\natexlab{a}}.

\bibitem[Li et~al.(2016)Li, Monroe, and Jurafsky]{li2016understanding}
Li, J., Monroe, W., and Jurafsky, D.
\newblock Understanding neural networks through representation erasure.
\newblock \emph{arXiv preprint arXiv:1612.08220}, 2016.

\bibitem[Li et~al.(2023{\natexlab{a}})Li, Li, Savarese, and Hoi]{li2023blip}
Li, J., Li, D., Savarese, S., and Hoi, S.
\newblock Blip-2: Bootstrapping language-image pre-training with frozen image encoders and large language models.
\newblock In \emph{International conference on machine learning}, pp.\  19730--19742. PMLR, 2023{\natexlab{a}}.

\bibitem[Li et~al.(2025{\natexlab{b}})Li, Wu, Jin, Chen, Ji, Sun, Cao, and Ji]{li2025mihbench}
Li, J., Wu, M., Jin, Z., Chen, H., Ji, J., Sun, X., Cao, L., and Ji, R.
\newblock Mihbench: Benchmarking and mitigating multi-image hallucinations in multimodal large language models.
\newblock \emph{arXiv preprint arXiv:2508.00726}, 2025{\natexlab{b}}.

\bibitem[Li et~al.(2023{\natexlab{b}})Li, Du, Zhou, Wang, Zhao, and Wen]{li-etal-2023-evaluating}
Li, Y., Du, Y., Zhou, K., Wang, J., Zhao, X., and Wen, J.-R.
\newblock Evaluating object hallucination in large vision-language models.
\newblock In \emph{Proceedings of the 2023 Conference on Empirical Methods in Natural Language Processing}, pp.\  292--305, 2023{\natexlab{b}}.

\bibitem[Lin et~al.(2014)Lin, Maire, Belongie, Hays, Perona, Ramanan, Doll{\'a}r, and Zitnick]{10.1007/978-3-319-10602-1_48}
Lin, T.-Y., Maire, M., Belongie, S., Hays, J., Perona, P., Ramanan, D., Doll{\'a}r, P., and Zitnick, C.~L.
\newblock Microsoft coco: Common objects in context.
\newblock In \emph{Computer Vision -- ECCV 2014}, pp.\  740--755, 2014.

\bibitem[Liu et~al.(2023)Liu, Li, Wu, and Lee]{liu2023visual}
Liu, H., Li, C., Wu, Q., and Lee, Y.~J.
\newblock Visual instruction tuning.
\newblock \emph{Advances in neural information processing systems}, 36:\penalty0 34892--34916, 2023.

\bibitem[Liu et~al.(2024{\natexlab{a}})Liu, Li, Li, and Lee]{liu2024improved}
Liu, H., Li, C., Li, Y., and Lee, Y.~J.
\newblock Improved baselines with visual instruction tuning.
\newblock In \emph{Proceedings of the IEEE/CVF conference on computer vision and pattern recognition}, pp.\  26296--26306, 2024{\natexlab{a}}.

\bibitem[Liu et~al.(2024{\natexlab{b}})Liu, Lin, Hewitt, Paranjape, Bevilacqua, Petroni, and Liang]{liu-etal-2024-lost}
Liu, N.~F., Lin, K., Hewitt, J., Paranjape, A., Bevilacqua, M., Petroni, F., and Liang, P.
\newblock Lost in the middle: How language models use long contexts.
\newblock \emph{Transactions of the Association for Computational Linguistics}, pp.\  157--173, 2024{\natexlab{b}}.

\bibitem[Lu et~al.(2025)Lu, Wu, Zheng, Ma, Gong, Liu, Zhai, Cao, Shen, and Zha]{lu2025benchmarking}
Lu, F., Wu, W., Zheng, K., Ma, S., Gong, B., Liu, J., Zhai, W., Cao, Y., Shen, Y., and Zha, Z.-J.
\newblock Benchmarking large vision-language models via directed scene graph for comprehensive image captioning.
\newblock In \emph{Proceedings of the Computer Vision and Pattern Recognition Conference}, pp.\  19618--19627, 2025.

\bibitem[Lu et~al.(2022)Lu, Mishra, Xia, Qiu, Chang, Zhu, Tafjord, Clark, and Kalyan]{lu2022learn}
Lu, P., Mishra, S., Xia, T., Qiu, L., Chang, K.-W., Zhu, S.-C., Tafjord, O., Clark, P., and Kalyan, A.
\newblock Learn to explain: Multimodal reasoning via thought chains for science question answering.
\newblock \emph{Advances in Neural Information Processing Systems}, 35:\penalty0 2507--2521, 2022.

\bibitem[Peng et~al.(2024)Peng, Quesnelle, Fan, and Shippole]{peng2024yarn}
Peng, B., Quesnelle, J., Fan, H., and Shippole, E.
\newblock Ya{RN}: Efficient context window extension of large language models.
\newblock In \emph{The Twelfth International Conference on Learning Representations}, 2024.

\bibitem[Qi et~al.(2025)Qi, Liu, Tang, and Zhu]{qi2025beyond}
Qi, J., Liu, J., Tang, H., and Zhu, Z.
\newblock Beyond semantics: Rediscovering spatial awareness in vision-language models.
\newblock \emph{arXiv preprint arXiv:2503.17349}, 2025.

\bibitem[Radford et~al.(2021)Radford, Kim, Hallacy, Ramesh, Goh, Agarwal, Sastry, Askell, Mishkin, Clark, Krueger, and Sutskever]{pmlr-v139-radford21a}
Radford, A., Kim, J.~W., Hallacy, C., Ramesh, A., Goh, G., Agarwal, S., Sastry, G., Askell, A., Mishkin, P., Clark, J., Krueger, G., and Sutskever, I.
\newblock Learning transferable visual models from natural language supervision.
\newblock In \emph{Proceedings of the 38th International Conference on Machine Learning}, volume 139, pp.\  8748--8763. PMLR, 2021.

\bibitem[Ruggeri et~al.(2023)Ruggeri, Nozza, et~al.]{ruggeri2023multi}
Ruggeri, G., Nozza, D., et~al.
\newblock A multi-dimensional study on bias in vision-language models.
\newblock In \emph{Findings of the Association for Computational Linguistics: ACL 2023}. Association for Computational Linguistics, 2023.

\bibitem[Schuhmann et~al.(2022)Schuhmann, Beaumont, Vencu, Gordon, Wightman, Cherti, Coombes, Katta, Mullis, Wortsman, et~al.]{schuhmann2022laion}
Schuhmann, C., Beaumont, R., Vencu, R., Gordon, C., Wightman, R., Cherti, M., Coombes, T., Katta, A., Mullis, C., Wortsman, M., et~al.
\newblock Laion-5b: An open large-scale dataset for training next generation image-text models.
\newblock \emph{Advances in neural information processing systems}, 35:\penalty0 25278--25294, 2022.

\bibitem[Shi et~al.(2025)Shi, Ma, Liang, Diao, Ma, and Vosoughi]{shi-etal-2025-judging}
Shi, L., Ma, C., Liang, W., Diao, X., Ma, W., and Vosoughi, S.
\newblock Judging the judges: A systematic study of position bias in {LLM}-as-a-judge.
\newblock In \emph{Proceedings of the 14th International Joint Conference on Natural Language Processing and the 4th Conference of the Asia-Pacific Chapter of the Association for Computational Linguistics}, pp.\  292--314, 2025.

\bibitem[Si et~al.(2024)Si, Zhu, Lu, and Zhou]{si-etal-2024-denoising}
Si, J., Zhu, Y., Lu, W., and Zhou, D.
\newblock Denoising rationalization for multi-hop fact verification via multi-granular explainer.
\newblock In \emph{Findings of the Association for Computational Linguistics: EMNLP 2024}, pp.\  12593--12608, 2024.

\bibitem[Singh et~al.(2019)Singh, Natarajan, Shah, Jiang, Chen, Batra, Parikh, and Rohrbach]{Singh_2019_CVPR}
Singh, A., Natarajan, V., Shah, M., Jiang, Y., Chen, X., Batra, D., Parikh, D., and Rohrbach, M.
\newblock Towards vqa models that can read.
\newblock In \emph{Proceedings of the IEEE/CVF Conference on Computer Vision and Pattern Recognition (CVPR)}, June 2019.

\bibitem[Su et~al.(2024)Su, Ahmed, Lu, Pan, Bo, and Liu]{su2024roformer}
Su, J., Ahmed, M., Lu, Y., Pan, S., Bo, W., and Liu, Y.
\newblock Roformer: Enhanced transformer with rotary position embedding.
\newblock \emph{Neurocomputing}, 568:\penalty0 127063, 2024.

\bibitem[Sun et~al.(2024)Sun, Shen, Cao, Liu, Li, Shen, Gan, Gui, Wang, Yang, Keutzer, and Darrell]{sun-etal-2024-aligning}
Sun, Z., Shen, S., Cao, S., Liu, H., Li, C., Shen, Y., Gan, C., Gui, L., Wang, Y.-X., Yang, Y., Keutzer, K., and Darrell, T.
\newblock Aligning large multimodal models with factually augmented {RLHF}.
\newblock In \emph{Findings of the Association for Computational Linguistics: ACL 2024}, pp.\  13088--13110, 2024.

\bibitem[Team et~al.(2025)Team, Kamath, Ferret, Pathak, Vieillard, Merhej, Perrin, Matejovicova, Ram{\'e}, Rivi{\`e}re, et~al.]{team2025gemma}
Team, G., Kamath, A., Ferret, J., Pathak, S., Vieillard, N., Merhej, R., Perrin, S., Matejovicova, T., Ram{\'e}, A., Rivi{\`e}re, M., et~al.
\newblock Gemma 3 technical report.
\newblock \emph{arXiv preprint arXiv:2503.19786}, 2025.

\bibitem[Tian et~al.(2025)Tian, Zou, Yang, and Zhang]{tian2025identifying}
Tian, X., Zou, S., Yang, Z., and Zhang, J.
\newblock Identifying and mitigating position bias of multi-image vision-language models.
\newblock In \emph{Proceedings of the Computer Vision and Pattern Recognition Conference}, pp.\  10599--10609, 2025.

\bibitem[Touvron et~al.(2023)Touvron, Lavril, Izacard, Martinet, Lachaux, Lacroix, Rozi{\`e}re, Goyal, Hambro, Azhar, et~al.]{touvron2023llama}
Touvron, H., Lavril, T., Izacard, G., Martinet, X., Lachaux, M.-A., Lacroix, T., Rozi{\`e}re, B., Goyal, N., Hambro, E., Azhar, F., et~al.
\newblock Llama: Open and efficient foundation language models.
\newblock \emph{arXiv preprint arXiv:2302.13971}, 2023.

\bibitem[Wang et~al.(2024{\natexlab{a}})Wang, Cao, Zhang, Yuan, Shan, Chen, and Gao]{wang2024vlbiasbench}
Wang, S., Cao, X., Zhang, J., Yuan, Z., Shan, S., Chen, X., and Gao, W.
\newblock Vlbiasbench: A comprehensive benchmark for evaluating bias in large vision-language model.
\newblock \emph{arXiv preprint arXiv:2406.14194}, 2024{\natexlab{a}}.

\bibitem[Wang et~al.(2024{\natexlab{b}})Wang, Ren, Luo, Li, Yan, Chen, Wang, Li, Lu, Zhu, et~al.]{wang2024all}
Wang, W., Ren, Y., Luo, H., Li, T., Yan, C., Chen, Z., Wang, W., Li, Q., Lu, L., Zhu, X., et~al.
\newblock The all-seeing project v2: Towards general relation comprehension of the open world.
\newblock In \emph{European Conference on Computer Vision}, pp.\  471--490. Springer, 2024{\natexlab{b}}.

\bibitem[Wang et~al.(2025{\natexlab{a}})Wang, Wu, Zhang, Yan, Liu, Luo, and Fei]{wang2025multimodal}
Wang, Y., Wu, S., Zhang, Y., Yan, S., Liu, Z., Luo, J., and Fei, H.
\newblock Multimodal chain-of-thought reasoning: A comprehensive survey.
\newblock \emph{arXiv preprint arXiv:2503.12605}, 2025{\natexlab{a}}.

\bibitem[Wang et~al.(2025{\natexlab{b}})Wang, Zhang, Li, Huang, Han, Ji, Kakade, Peng, and Ji]{wang2025eliminating}
Wang, Z., Zhang, H., Li, X., Huang, K.-H., Han, C., Ji, S., Kakade, S.~M., Peng, H., and Ji, H.
\newblock Eliminating position bias of language models: A mechanistic approach.
\newblock In \emph{The Thirteenth International Conference on Learning Representations}, 2025{\natexlab{b}}.

\bibitem[Xing et~al.(2024)Xing, Li, Laptev, and Lu]{xing2024mitigating}
Xing, Y., Li, Y., Laptev, I., and Lu, S.
\newblock Mitigating object hallucination via concentric causal attention.
\newblock \emph{Advances in neural information processing systems}, 37:\penalty0 92012--92035, 2024.

\bibitem[Xu et~al.(2025)Xu, Shao, Zhang, Gao, Liu, Lei, Meng, Huang, Qiao, and Luo]{10769058}
Xu, P., Shao, W., Zhang, K., Gao, P., Liu, S., Lei, M., Meng, F., Huang, S., Qiao, Y., and Luo, P.
\newblock Lvlm-ehub: A comprehensive evaluation benchmark for large vision-language models.
\newblock \emph{IEEE Transactions on Pattern Analysis and Machine Intelligence}, 47\penalty0 (3):\penalty0 1877--1893, 2025.

\bibitem[Yu et~al.(2025)Yu, Jiang, Luo, Wu, Lin, Li, Yang, Huang, and Qiu]{yu-etal-2025-mitigate}
Yu, Y., Jiang, H., Luo, X., Wu, Q., Lin, C.-Y., Li, D., Yang, Y., Huang, Y., and Qiu, L.
\newblock Mitigate position bias in {LLM}s via scaling a single hidden states channel.
\newblock In \emph{Findings of the Association for Computational Linguistics: ACL 2025}, pp.\  6092--6111, 2025.

\bibitem[Yue et~al.(2025)Yue, Zheng, Ni, Wang, Zhang, Tong, Sun, Yu, Zhang, Sun, Su, Chen, and Neubig]{yue-etal-2025-mmmu}
Yue, X., Zheng, T., Ni, Y., Wang, Y., Zhang, K., Tong, S., Sun, Y., Yu, B., Zhang, G., Sun, H., Su, Y., Chen, W., and Neubig, G.
\newblock {MMMU}-pro: A more robust multi-discipline multimodal understanding benchmark.
\newblock In \emph{Proceedings of the 63rd Annual Meeting of the Association for Computational Linguistics (Volume 1: Long Papers)}, pp.\  15134--15186, 2025.

\bibitem[Zhang et~al.(2025{\natexlab{a}})Zhang, Yang, Inala, Singh, Gao, Su, and Wang]{zhang2025towards}
Zhang, K., Yang, J., Inala, J.~P., Singh, C., Gao, J., Su, Y., and Wang, C.
\newblock Towards understanding graphical perception in large multimodal models.
\newblock \emph{arXiv preprint arXiv:2503.10857}, 2025{\natexlab{a}}.

\bibitem[Zhang et~al.(2025{\natexlab{b}})Zhang, Ma, Hou, Bai, Chen, Xiang, Yu, and Zhang]{zhang2025evaluating}
Zhang, Y., Ma, J., Hou, Y., Bai, X., Chen, K., Xiang, Y., Yu, J., and Zhang, M.
\newblock Evaluating and steering modality preferences in multimodal large language model.
\newblock \emph{arXiv preprint arXiv:2505.20977}, 2025{\natexlab{b}}.

\bibitem[Zheng et~al.(2023)Zheng, Chiang, Sheng, Zhuang, Wu, Zhuang, Lin, Li, Li, Xing, et~al.]{zheng2023judging}
Zheng, L., Chiang, W.-L., Sheng, Y., Zhuang, S., Wu, Z., Zhuang, Y., Lin, Z., Li, Z., Li, D., Xing, E., et~al.
\newblock Judging llm-as-a-judge with mt-bench and chatbot arena.
\newblock \emph{Advances in neural information processing systems}, 36:\penalty0 46595--46623, 2023.

\bibitem[Zhu et~al.(2023)Zhu, Chen, Shen, Li, and Elhoseiny]{zhu2023minigpt}
Zhu, D., Chen, J., Shen, X., Li, X., and Elhoseiny, M.
\newblock Minigpt-4: Enhancing vision-language understanding with advanced large language models.
\newblock \emph{arXiv preprint arXiv:2304.10592}, 2023.

\bibitem[Zhu et~al.(2025{\natexlab{a}})Zhu, Wang, Chen, Liu, Ye, Gu, Tian, Duan, Su, Shao, et~al.]{zhu2025internvl3}
Zhu, J., Wang, W., Chen, Z., Liu, Z., Ye, S., Gu, L., Tian, H., Duan, Y., Su, W., Shao, J., et~al.
\newblock Internvl3: Exploring advanced training and test-time recipes for open-source multimodal models.
\newblock \emph{arXiv preprint arXiv:2504.10479}, 2025{\natexlab{a}}.

\bibitem[Zhu et~al.(2024)Zhu, Wang, Zhou, Wang, Chen, Wang, Yang, Ye, Zhang, Gong, and Xie]{zhu24PromptRobust}
Zhu, K., Wang, J., Zhou, J., Wang, Z., Chen, H., Wang, Y., Yang, L., Ye, W., Zhang, Y., Gong, N., and Xie, X.
\newblock Promptrobust: Towards evaluating the robustness of large language models on adversarial prompts.
\newblock In \emph{Proceedings of the 1st ACM Workshop on Large AI Systems and Models with Privacy and Safety Analysis}, pp.\  57–68, 2024.

\bibitem[Zhu et~al.(2025{\natexlab{b}})Zhu, Bai, Chen, Xiang, Yu, and Zhang]{zhu-etal-2025-benchmarking}
Zhu, Y., Bai, X., Chen, K., Xiang, Y., Yu, J., and Zhang, M.
\newblock Benchmarking and improving large vision-language models for fundamental visual graph understanding and reasoning.
\newblock In \emph{Proceedings of the 63rd Annual Meeting of the Association for Computational Linguistics (Volume 1: Long Papers)}, pp.\  30678--30701, 2025{\natexlab{b}}.

\bibitem[Zhu et~al.(2025{\natexlab{c}})Zhu, Tao, Dong, and Xu]{zhu2025mitigating}
Zhu, Y., Tao, L., Dong, M., and Xu, C.
\newblock Mitigating object hallucinations in large vision-language models via attention calibration.
\newblock \emph{arXiv preprint arXiv:2502.01969}, 2025{\natexlab{c}}.

\end{thebibliography}
\bibliographystyle{icml2026}

\newpage
\appendix
\onecolumn

\section*{Limitations}
This paper presents a systematic investigation of spatial bias in LVLMs. Although providing new insights and solutions, our work suffers from two limitations.
First, our probing experiments focus primarily on a $3\times 3$ spatial grid, which may not fully capture more fine-grained or irregular spatial variations. Second, due to computational constraints, we limit our evaluation to medium-scale LVLMs and do not include very large models such as Qwen3-VL-235B-A22B. Investigating whether the identified spatial bias persists or evolves in larger-scale models remains an important direction for future work.

\section{Details of Probe Task}
\label{app:probe_task}
\begin{figure}[ht]
    \centering
    \includegraphics[width=\linewidth]{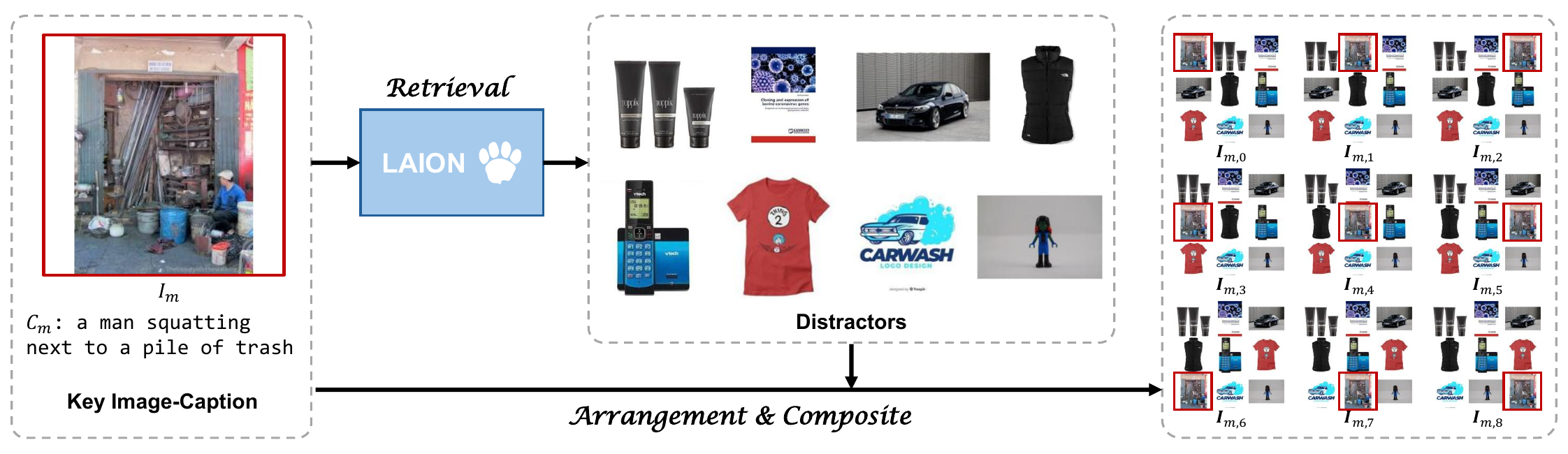}
    \caption{Workflow illustration on how we synthesize composite images in probe dataset, where the key image features red borders not present in our experiments.}
    \label{fig:probe_dataset}
\end{figure}

For each key image $I_m$ and caption $C_m$ randomly selected from LAION, we first generate 9 corresponding composite images $\{\boldsymbol{I}_{m,n}\}^8_{n=0}$ according to the workflow depicted in Figure~\ref{fig:probe_dataset}. Then we fed each composite image $\boldsymbol{I}_{m,n}$ alongside the question $Q_m$ format as following into the LVLMs to test their spatial robustness.

\begin{tcolorbox}[
    title=Question $Q_m$, 
    colback=gray!5!white, 
    colframe=gray!75!black, 
    fonttitle=\bfseries 
]
    Determine if there is a sub-image in the given image that matches the text following.
    
    Text: \{$C_m$\}
    
    The answer should only contain `Yes' or `No', without reasoning process.
\end{tcolorbox}

\begin{figure}[ht]
    \centering
    \includegraphics[width=\linewidth]{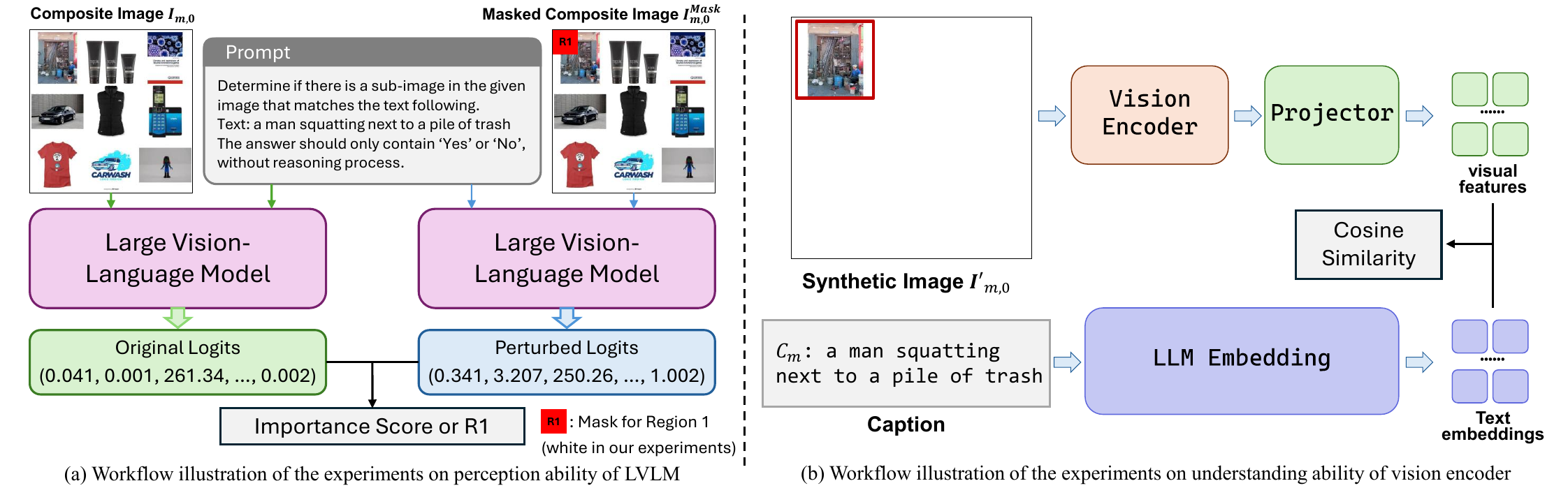}
    \caption{Workflow illustration on our analysis experiments.}
    \label{fig:explore}
\end{figure}

\section{Experiments on Different Image Resolutions}
\label{app:image_resolution}
\begin{figure}[ht]
    \centering
    \includegraphics[width=0.8\linewidth]{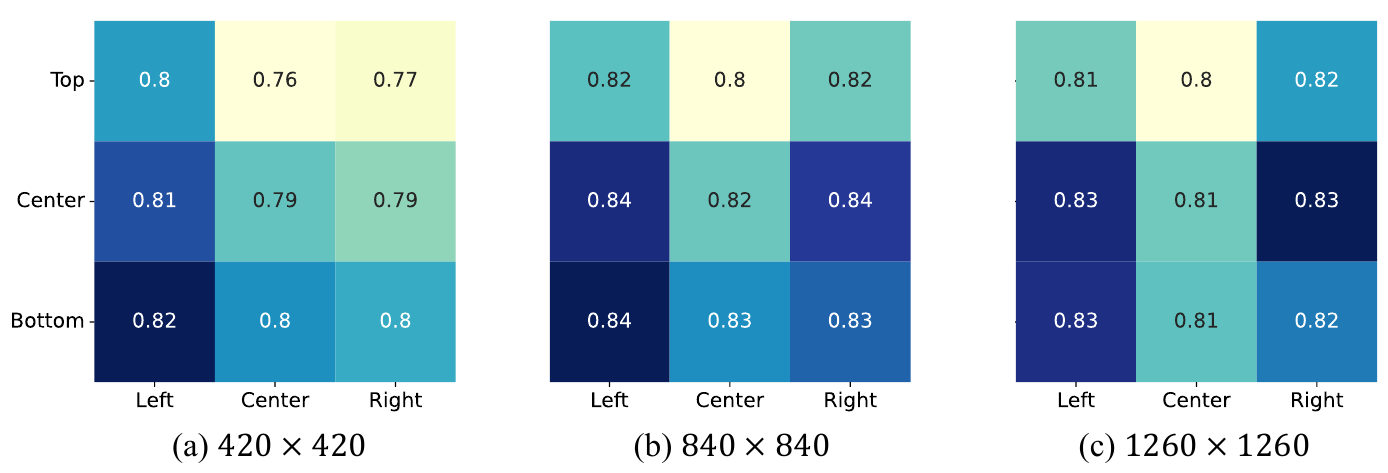}
    \caption{Results on our probe dataset with different resolution of the input image.}
    \label{fig:resolution}
\end{figure}

To further explore the relationship between spatial bias and RoPE, we evaluated the performance of Qwen2.5-VL on our probe datasets featuring input images of varying resolutions. The results are shown in Figure~\ref{fig:resolution}. It can be found that although higher resolutions introduce a larger number of visual tokens, the overall spatial bias patterns remain largely consistent across resolutions. Notably, we observe that lower-resolution inputs even exhibit more severe bias at positions farther away from the text tokens compared to higher-resolution settings. This counterintuitive behavior suggests that spatial bias does not scale with token distance as would be expected from RoPE-induced long-term decay.

\begin{figure}[ht]
    \centering
    \subfigure[Qwen2.5-VL]{
        \includegraphics[width=0.32\textwidth]{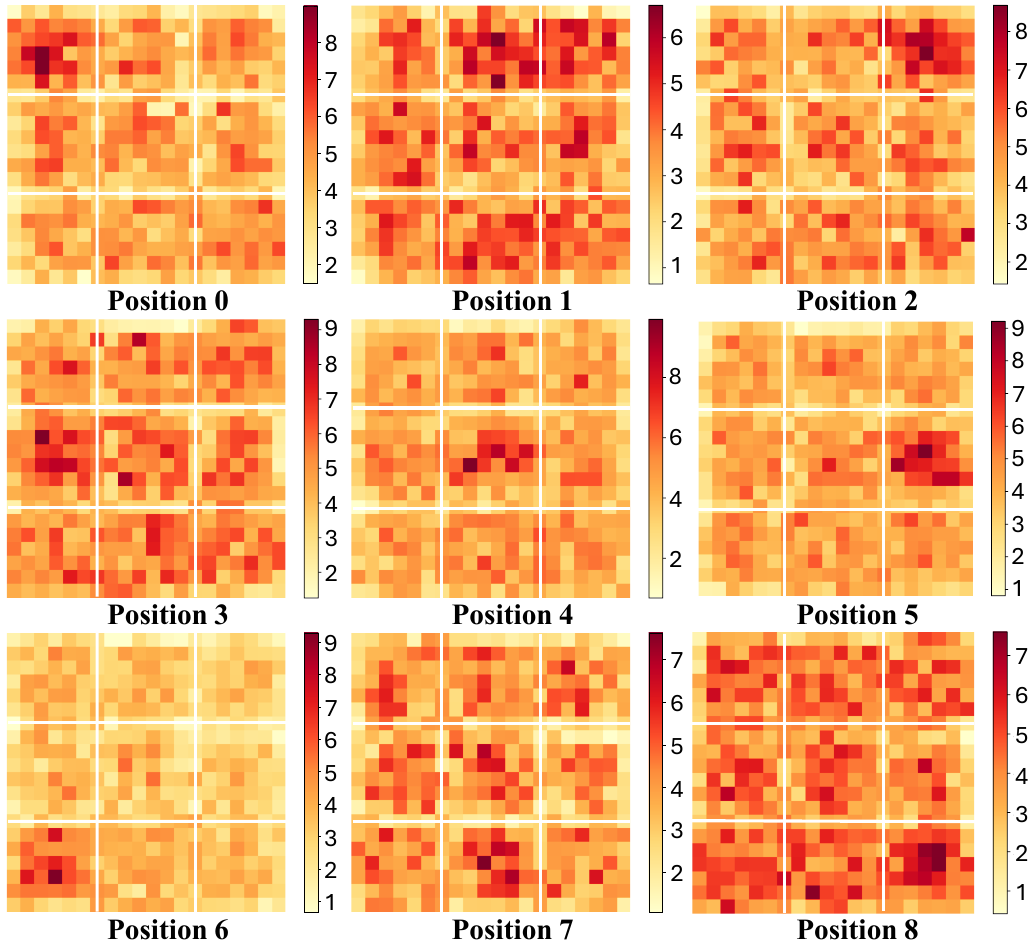}
    }
    \hspace{-0.2cm}
    \hfill
    \subfigure[Qwen3-VL]{
        \includegraphics[width=0.32\textwidth]{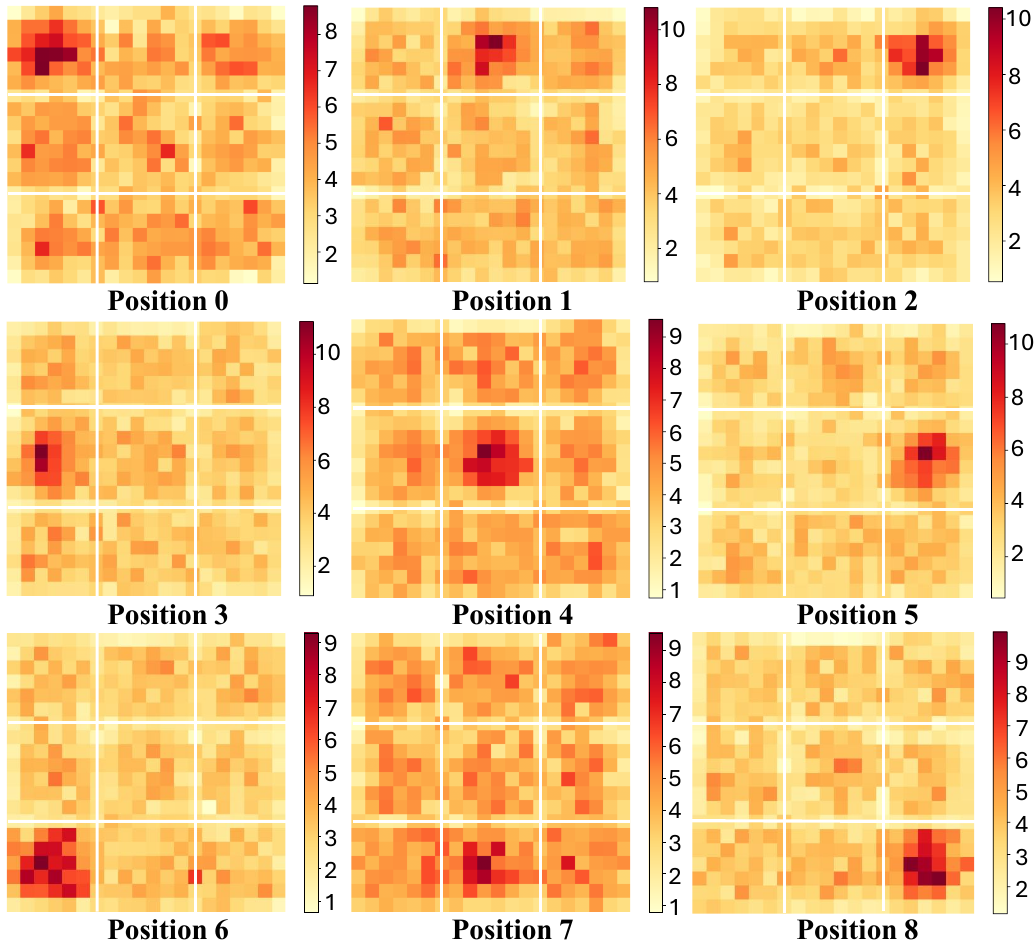}
    }
    \hfill
    \subfigure[LLaVA-v1.6]{
        \includegraphics[width=0.32\textwidth]{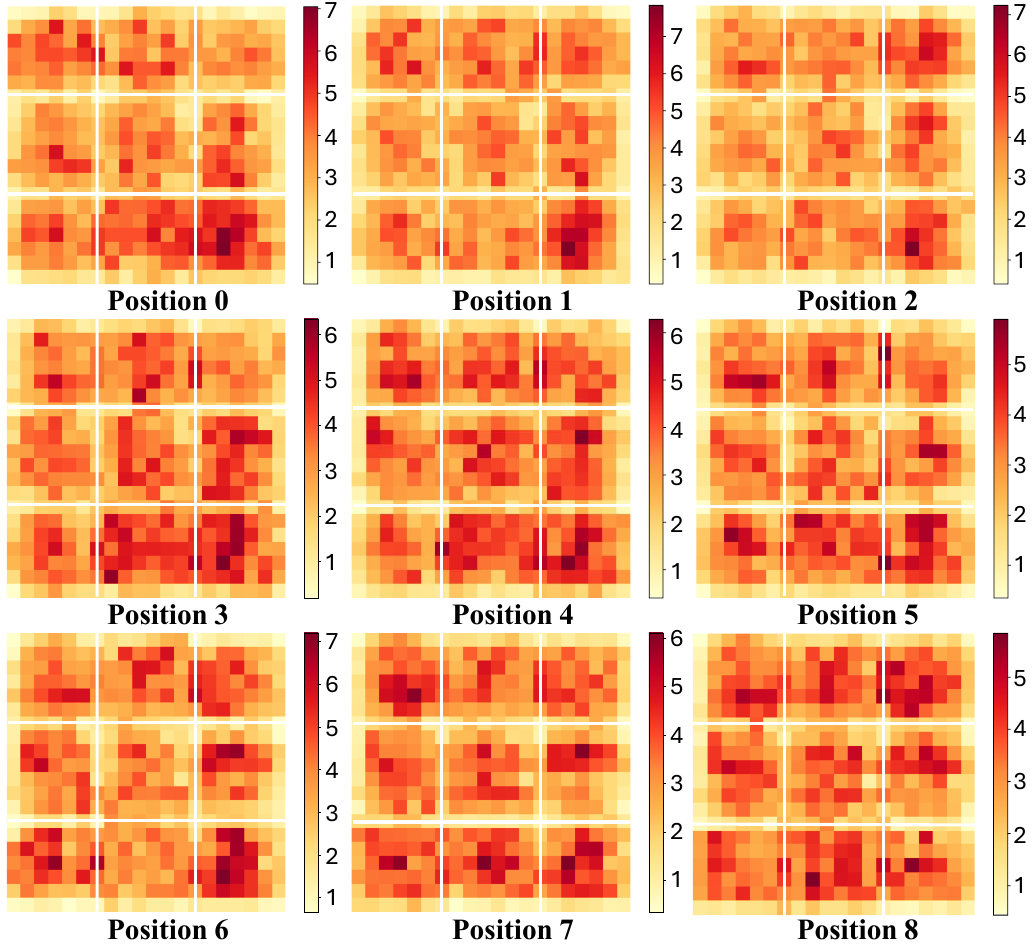}
    }
    \caption{More results on perception ability across each \textit{Position} $n$ of key image. Darker regions indicate higher importance scores, where logits change more significantly before and after masking.}
    \label{fig:mask_results_more}
\end{figure}

\section{Mutual Information Analysis of AGCI}
\label{app:mi_agci}

We state a simple sufficient condition under which AGCI is guaranteed to not reduce the information available for text prediction.

\begin{proposition}
\label{prop:mi_agci}
Let $\{t_j\}_{j=1}^{N}$ denote the (pre-softmax) hidden representations of the $N$ text tokens.
Let $r\triangleq \psi(t_1,\dots,t_{j^\ast})$ be the text-side readout used by the model to produce the final prediction $y$.
For a standard causal LLM, one can take $r=t_{j^\ast}$ (the hidden state at the prediction position), which already summarizes all previous tokens.
Assume that for each $i$, the injected token is computed as $\tilde{v}_i' = \phi_i(\tilde{v}_i,\tilde{g})$ where $\tilde{g}$ is the global context computed from the image tokens, and $\phi_i$ is injective in $\tilde{v}_i$ given $(\tilde{g},w_i)$ (i.e., $\tilde{v}_i$ can be recovered from $(\tilde{v}_i',\tilde{g},w_i)$).
Then
\begin{equation}
    I(\tilde{v}_i',\tilde{g}; r) \ge I(\tilde{v}_i; r) \ge I(\tilde{v}_i; y).
\end{equation}
\end{proposition}
\begin{proof}
Since $y$ is a deterministic function of $r$ (via a fixed output head), the data processing inequality gives
\begin{equation}
    I(\tilde{v}_i; y) \le I(\tilde{v}_i; r).
\end{equation}
Because $(\tilde{v}_i',\tilde{g})$ is a deterministic function of $(\tilde{v}_i,\tilde{g})$, we have
\begin{equation}
    I(\tilde{v}_i; r) \le I(\tilde{v}_i,\tilde{g}; r).
\end{equation}
By the injectivity assumption, $(\tilde{v}_i,\tilde{g})$ is a deterministic function of $(\tilde{v}_i',\tilde{g},w_i)$, and thus (another application of data processing)
\begin{equation}
    I(\tilde{v}_i,\tilde{g}; r) \le I(\tilde{v}_i',\tilde{g}; r).
\end{equation}
Combining these inequalities yields the claim.
\end{proof}

Consequently, once the model has access to (an estimate of) the global context $\tilde{g}$, any preferred spatial token can provide at least as much predictive information as the original token, which helps mitigate spatial bias caused by uneven token preference.

\section{Experimental Settings}
\label{app:experimental_settings}
\subsection{Downstream Datasets}
\paragraph{MMMU-Pro} MMMU-Pro~\citep{yue-etal-2025-mmmu} is an enhanced multimodal benchmark designed to  challenge and evaluate multimodal models with tasks demanding college-level subject knowledge and complex reasoning. It contains 1.73K meticulously collected multimodal questions from college exams, quizzes, and textbooks, covering six core disciplines: Art \& Design, Business, Science, Health \& Medicine, Humanities \& Social Science, and Tech \& Engineering. To facilitate statistical analysis, we evaluate each LVLM under 4-option and 10-option settings, respectively, and ask them to answer questions directly. 

\paragraph{ScienceQA}
ScienceQA~\citep{lu2022learn} is a large-scale multi-choice dataset collected from elementary and high school science curricula, and contains 21,208 multimodal science questions with explanations and features rich domain diversity. We assess each LVLM on the test set of ScienceQA with 4,241 samples.

\paragraph{CRPE}
CRPE is a benchmark designed to quantitatively evaluate the object recognition and relation comprehension ability of LVLMs. The evaluation is formulated as single-choice questions.
Following previous work~\citep{bai2025qwen2}, we evaluate LVLMs on relation comprehension ability instead of recognition with CRPE.

\paragraph{HallusionBench}
HallusionBench is a comprehensive benchmark designed for the evaluation of image-context reasoning, which comprises 346 images paired with 1129 questions, all meticulously crafted by human experts.

\paragraph{TextVQA} TextVQA requires models to read and reason about text in an image to answer questions based on them, containing contains 45,336 questions on 28,408 images. We test each LVLM on the validation set with 5,000 smaples and evaluate the results through the scripts provided by~\citet{liu2023visual}\footnote{\url{https://github.com/haotian-liu/LLaVA/blob/main/scripts/v1_5/eval/textvqa.sh}}.

\paragraph{POPE} Polling-based Object Probing Evaluation (POPE), a simple yet effective approach for evaluating object hallucination in LVLMs~\citep{li-etal-2023-evaluating}. It formulates the evaluation as a binary classification task that prompts LVLMs to output ``Yes'' or ``No'' and devise three object sampling strategies, including random sampling, popular sampling and adversarial sampling. We evaluate each LVLM with POPE bulit on the validation set of MSCOCO, containing 500 images with 9,000 questions in total.



\subsection{Implementation Details}
\label{app:implementation}
For each LVLM, we randomly select 10,000 samples from the LLaVA-v1.5 instruction-tuning data\footnote{\url{https://huggingface.co/datasets/liuhaotian/LLaVA-Instruct-150K/blob/main/llava_v1_5_mix665k.json}} and fine-tune the models with LoRA~\citep{hu2022lora} using the LLaMA-Factory library\footnote{\url{https://github.com/hiyouga/LLaMA-Factory}}. The hyperparameters used for
training are shown in Table~\ref{tab:hyperparameters}. 
To better adapt LLaVA-v1.5 to AGCI, we retrain it from scratch using the same two stages as~\citet{xing2024mitigating}, and the hyperparameters are presented in Table~\ref{tab:llava_hp}
All the experiments are finished on 4 NVIDIA H20 GPUs with 96GB memory.
\begin{table}[ht]
\centering
\small
\caption{Hyperparameters for LoRA finetuning.}
\label{tab:hyperparameters}
 \renewcommand\arraystretch{1.05}
{
\begin{tabular}{lccccc}
\toprule
\textbf{Model}      & \textbf{Lora\_rank}  & \textbf{Lora\_alpha} & \textbf{Global Batch Size} & \textbf{Learning rate} & \textbf{Epoch}       \\
\midrule
\textbf{Gemma3n-E4B}     & 8                    & 16                   & 32                        & 1.0e-05                & 2                    \\
\textbf{LLaVA-v1.6-7B} & 8                    & 16                   & 64                         & 1.0e-04                & 1                    \\
\textbf{InternVL3-8B} & 8                    & 16                   & 16                         & 1.0e-04                & 2                    \\
\textbf{Qwen3-VL-8B} & 8                    & 16                   & 64                        & 1.0e-04                & 1                    \\
\textbf{Qwen2.5-VL-7B} & 8                    & 16                   & 64                        & 1.0e-04                & 1                    \\
\textbf{Qwen2.5-VL-32B} & 8                    & 16                   & 64                        & 1.0e-06                & 1                    \\
\textbf{Qwen2.5-VL-72B} & 8                    & 16                   & 64                        & 1.0e-04               & 1                    \\
\bottomrule
\end{tabular}
}
\end{table}

\begin{table}[ht]
\centering
\small
\caption{Hyperparameters for training LLaVA-v1.5 from scratch.}
\label{tab:llava_hp}
\begin{tabular}{lccccc}
\toprule
\textbf{Stage}  & \textbf{Global Batch Size} & \textbf{Learning rate} & \textbf{Epochs} & \textbf{Max length} & \textbf{Weight decay} \\
\midrule
Pre-training    & 256                        & 1.0e-03                & 1               & 2048                & 0                     \\
Instruct-tuning & 128                        & 2.0e-05                & 1               & 2048                & 0                    \\
\bottomrule
\end{tabular}
\end{table}

\begin{table}[ht]
\tiny
\centering
\caption{Results on POPE dataset. The best results are \textbf{bolded}}
\label{tab:pope1}
\resizebox{0.75\linewidth}{!}{
 \renewcommand\arraystretch{1.1}
\begin{tabular}{lcccccccc}
\toprule
\multirow{2.2}{*}{\textbf{Models}} & \multicolumn{2}{c}{\textbf{Adversarial}} & \multicolumn{2}{c}{\textbf{Popular}} & \multicolumn{2}{c}{\textbf{Random}} & \multicolumn{2}{c}{\textbf{Average}} \\ \cmidrule(lr){2-3}\cmidrule(lr){4-5}\cmidrule(lr){6-7}\cmidrule(lr){8-9}
                                 & \textbf{Acc}        & \textbf{F1}        & \textbf{Acc}      & \textbf{F1}      & \textbf{Acc}     & \textbf{F1}      & \textbf{Acc}      & \textbf{F1}      \\ \midrule
\textbf{Gemma3n}                 & 86.77               & 85.61              & 87.90             & 86.69            & 88.93            & 87.69            & 87.87             & 86.66            \\
\textbf{Gemma3n-AGCI}      & \textbf{87.07}      & \textbf{86.11}     & \textbf{88.37}    & \textbf{87.32}   & \textbf{89.40}   & \textbf{88.32}   & \textbf{88.28}    & \textbf{87.25}   \\ \hdashline
\textbf{LLaVA-v1.6}              & 87.10               & 86.55              & 88.70             & 88.02            & 90.80            & 90.02            & 88.87             & 88.20            \\
\textbf{LLaVA-v1.6-AGCI}   & \textbf{87.17}      & \textbf{86.79}     & \textbf{89.23}    & \textbf{88.69}   & \textbf{91.40}   & \textbf{90.75}   & \textbf{89.27}    & \textbf{88.74}   \\\hdashline
\textbf{InternVL3}               & 86.43               & 85.43              & 88.00             & 86.90            & 89.20            & 88.05            & 87.88             & 86.79            \\
\textbf{InternVL3-AGCI}    & \textbf{86.57}      & \textbf{85.70}     & \textbf{88.13}    & \textbf{87.16}   & \textbf{89.57}   & \textbf{88.53}   & \textbf{88.09}    & \textbf{87.13}   \\\hdashline
\textbf{Qwen3-VL-8B}             & 82.89               & 83.51              & 84.10             & 84.49            & 86.54            & 86.85            & 84.51             & 84.84            \\
\textbf{Qwen3-VL-AGCI}     & \textbf{84.67}      & \textbf{84.96}     & \textbf{86.57}    & \textbf{86.58}   & \textbf{89.63}   & \textbf{89.32}   & \textbf{86.96}    & \textbf{86.95}   \\\hdashline
\textbf{Qwen2.5-VL-7B}           & 87.77               & 88.12              & 90.67             & 90.67            & \textbf{94.13}   & 93.93            & 90.86             & \textbf{90.91}   \\
\textbf{Qwen2.5-VL-AGCI}   & \textbf{89.00}      & \textbf{88.88}     & \textbf{90.73}    & \textbf{90.47}   & 93.17            & \textbf{92.79}   & \textbf{90.97}    & 90.72            \\\hdashline
\textbf{LLaVA-v1.5}              & 78.96               & 77.57              & 81.88    & 80.06   & 83.29            & 81.33            & 81.38             & 79.65            \\
\textbf{LLaVA-v1.5-VCD}          &   80.88    & 81.33             & 85.38            & 85.06            & 87.73            & 87.16           & 84.66             & 84.52            \\
\textbf{LLaVA-v1.5-RLHF}          & 82.60      & 80.88              & 83.90             & 82.05            & 85.90           & 83.92            & 84.13             & 82.28            \\
\textbf{LLaVA-v1.5-CCA}          & \textbf{85.67}      & 84.42              & \textbf{86.87}             & 85.54            & 88.03           & 86.65            & 86.86             & 85.54            \\
\textbf{LLaVA-v1.5-AGCI}   & 85.50               & \textbf{84.51}     & \textbf{86.87}            & \textbf{85.76}            & \textbf{88.57}   & \textbf{87.37}   & \textbf{86.98}    & \textbf{85.88}  \\ \bottomrule
\end{tabular}}
\end{table}


\section{Case Study}
\label{app:case_study}

\begin{figure}[ht]
    \centering
    \includegraphics[width=0.6\linewidth]{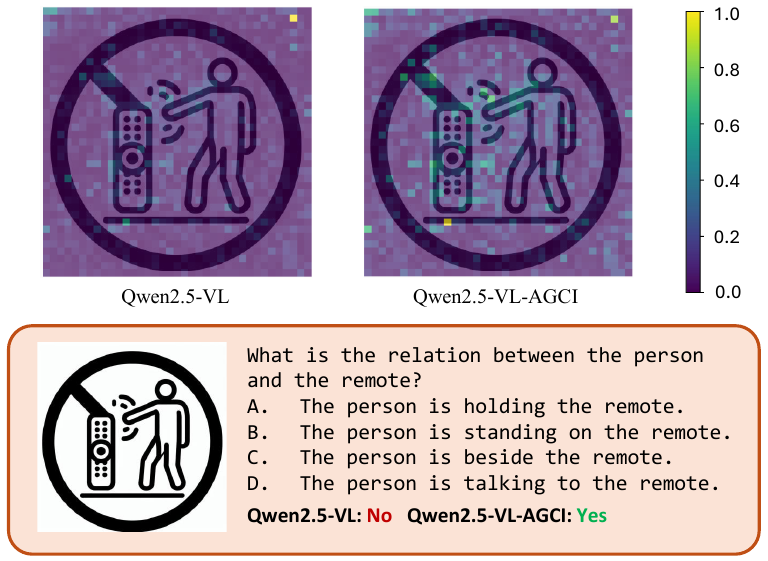}
    \caption{An example from CRPE benchmark.}
    \label{fig:case_study_1}
\end{figure}

We further randomly select one example CRPE for demonstration. The results are shown in Figures~\ref{fig:case_study_1}.

\end{document}